\documentclass{article}
\usepackage{microtype}
\usepackage{graphicx}
\usepackage{subcaption}
\usepackage{booktabs}
\usepackage{hyperref}

\usepackage{multirow}
\usepackage[table,xcdraw]{xcolor}
\usepackage{amsmath}
\usepackage{amssymb}
\usepackage{xcolor}
\usepackage[table,xcdraw]{xcolor}
\usepackage[normalem]{ulem}
\useunder{\uline}{\ul}{}
\usepackage{booktabs}
\usepackage{xspace}
\newcommand{\EXaMCaP}{\textsc{EXaMCaP}\xspace}
\newcommand{\EXaM}{\textsc{EXaM}\xspace}
\usepackage[preprint]{icml2026}
\usepackage{amsmath}
\usepackage{amssymb}
\usepackage{mathtools}
\usepackage{amsthm}

\usepackage[capitalize,noabbrev]{cleveref}

\theoremstyle{plain}

\theoremstyle{definition}

\theoremstyle{remark}

\usepackage[textsize=tiny]{todonotes}

\icmltitlerunning{Subset Selection with Entropy Gain Maximization for Probing Capability Gains of Large Chart Understanding Training Sets}

\begin{document}

\twocolumn[
  \icmltitle{\EXaMCaP: Subset Selection with Entropy Gain Maximization for Probing Capability Gains of Large Chart Understanding Training Sets}

  \icmlsetsymbol{corresponding}{\dag}

  \begin{icmlauthorlist}
    \icmlauthor{Jiapeng Liu}{iie,ucas}
    \icmlauthor{Liang Li}{iie,corresponding}
    \icmlauthor{Bing Li}{iie}
    \icmlauthor{Peng Fu}{iie}
    \icmlauthor{Xiyan Gao}{iie}
    \icmlauthor{Chengyang Fang}{jxcj}
    \icmlauthor{Xiaoshuai Hao}{xiaomi}
    \icmlauthor{Can Ma}{iie}
  \end{icmlauthorlist}

  \icmlaffiliation{iie}{Institute of Information Engineering, Chinese Academy of Sciences, Beijing, China}
  \icmlaffiliation{ucas}{School of Cyberspace Security, University of Chinese Academy of Sciences, Beijing, China}
  \icmlaffiliation{jxcj}{School of Computer and Artificial Intelligence, Jiangxi University of Finance and Economics, Jiangxi, China}
  \icmlaffiliation{xiaomi}{Xiaomi EV, Xiaomi Campus, Anningzhuang Road, Beijing, China}

  \icmlcorrespondingauthor{Liang Li}{liliang@iie.ac.cn}
 
  \icmlkeywords{chart understanding, data selection}

  \vskip 0.3in
]

\printAffiliationsAndNotice{}
\begin{abstract}
 
Recent works focus on synthesizing Chart Understanding (ChartU) training sets to inject advanced chart knowledge into Multimodal Large Language Models (MLLMs), where the sufficiency of the knowledge is typically verified by quantifying capability gains via the fine-tune-then-evaluate paradigm. 
However, full-set fine-tuning MLLMs to assess such gains incurs significant time costs, hindering the iterative refinement cycles of the ChartU dataset. 
Reviewing the ChartU dataset synthesis and data selection domains, we find that subsets can potentially probe the MLLMs' capability gains from full-set fine-tuning. 
Given that data diversity is vital for boosting MLLMs' performance and entropy reflects this feature, we propose \EXaMCaP, which uses entropy gain maximization to select a subset. 
To obtain a high-diversity subset, \EXaMCaP chooses the maximum-entropy subset from the large ChartU dataset. 
As enumerating all possible subsets is impractical, \EXaMCaP iteratively selects samples to maximize the gain in set entropy relative to the current set, approximating the maximum-entropy subset of the full dataset. 
Experiments show that \EXaMCaP outperforms baselines in probing the capability gains of the ChartU training set, along with its strong effectiveness across diverse subset sizes and compatibility with various MLLM architectures.

\end{abstract}

\section{Introduction}

Chart Understanding (ChartU) aims to enable Multimodal Large Language Models (MLLMs) to accurately answer both descriptive questions that focus on recognizing basic chart elements, and reasoning questions that require analytical abilities~\cite{yang2025effective}. 
Advanced general MLLMs~\cite{chen2024far, chen2024internvl, bai2025qwen2} gain the abilities to process visual cues via well-designed architectures and extensive pre-training, enabling chart reasoning through prompt engineering.
However, most of them are less effective for more realistic and complex charts~\cite{wang2024charxiv, xia2025chartx} as they suffer from insufficient ability to perceive or reason about complex and abstract visual chart elements~\cite{liu2024mmc}, or inadequate coverage of chart types in real-world scenarios~\cite{xu2025chartm}.

\begin{figure}[t]
  \includegraphics[width=\columnwidth]{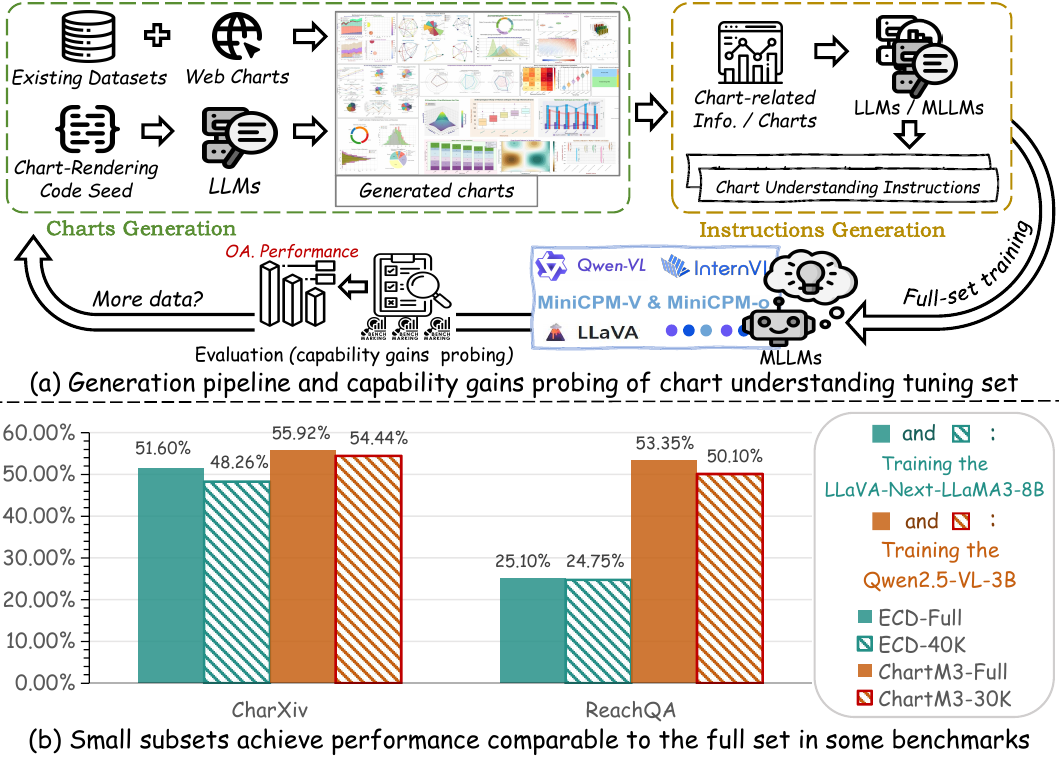}
   \caption{(a) General training set generation pipeline and dataset capability gain probing with full-set finetuning, (b) performance of subsets is comparable to the full sets.}
  \label{fig:exam_intro}
\end{figure}

Improving MLLMs' understanding largely depends on knowledge provided by the training data~\cite{chen2024far}. 
To further accelerate the evolution of MLLMs' ChartU capabilities, recent research efforts~\cite{han2023chartllama, masry2024chartinstruct,masry2025chartgemma,yang2025effective,xu2025chartm} focus on automatically generating large-scale ChartU datasets for visual instruction tuning. 
To assess the capability gains conferred by a generated ChartU dataset, a common paradigm is "fine-tune-then-evaluate": diverse MLLMs are fine-tuned on the complete dataset, and their newly acquired knowledge is verified through rigorous evaluation on multiple downstream test sets. 
In this way, researchers can quantify the improvement in the model ChartU's capabilities brought by the synthetic dataset. 
This helps decide whether to scale the data further or explore new directions for data generation.

However, evaluating the synthetic dataset's contribution to ChartU's capabilities—by performing full-set fine-tuning across diverse continuous iterations of MLLMs~\cite{wang2024qwen2, bai2025qwen2, zhu2025internvl3}—creates a severe time and computation bottleneck, stalling the iterative refinement cycle.
We review recent studies in ChartU and Data Selection domains and note that it is potential to utilize a subset to verify or probe the capabilities gains of ChartU that the full set can bring. 
Firstly, we summarize the experiments of existing works on ChartU training set generation~\cite{yang2025effective, xu2025chartm} and find that training on a small subset yields performance on par with full-set training across some benchmarks as shown in Figure~\ref{fig:exam_intro} (b).
Meanwhile, we note that for a general visual instruction tuning dataset, recent data selection methods~\cite{lee2024concept, bi2025prism} indicate that training on a subset can yield performance comparable to or even surpassing that of the full set.

Nevertheless, these general data selection methods are challenging to apply directly to curate a representative subset from a large-scale chart understanding dataset for evaluating 
the capability gains of the full dataset.
Specifically, some common approaches~\cite{yu2025mastering, safaei2025filter} 
rely on task-split markers to design their selection strategies, whereas the ChartU training set inherently encompasses multiple reasoning tasks without explicit fine-grained task division. 
As for gradient-based methods~\cite{liu2025less, wu2024icons}, their iterative gradient computations incur prohibitively high overhead—often comparable to the cost of full-set training itself.

Selecting a succinct subset that captures the multifaceted knowledge domains of the full ChartU dataset is a critical challenge for efficiently evaluating fine-tuning gains. 
This paper is motivated by simple yet profound observations: (1) the diversity of the training data is vital to improving MLLMs' performance due to broad knowledge coverage; and (2) entropy serves as a metric for quantifying system uncertainty and inherently guides distribution diversity~\cite{yang2025effective, friedman2022vendi}.
Building upon these insights, we introduce \EXaMCaP, a novel framework that employs \uline{E}ntropy \uline{Ga}in \uline{M}aximization as the sampling criterion to identify the most informative subset 
for probing the capability gains of the ChartU training set.
The core of \EXaMCaP lies in the entropy gain maximization sampling, i.e., selecting a highly diverse subset, as measured by entropy, to cover comprehensive ChartU knowledge.
Specifically, EXaMCaP iteratively selects samples that maximize entropy gain relative to the current set. 
This entropy is computed from an eigenvalue distribution derived from the similarity matrix. 
Such an iterative approach is adopted because exhaustive subset enumeration is computationally impractical, and the global probability distribution of these subsets remains unknown.
Furthermore, considering the high computational cost of greedy selection on large datasets, it is restricted to clusters from prior partitioning.
This helps to reduce the search space while ensuring global core feature coverage via coarse-grained sample partitioning.
Additionally, \EXaMCaP incorporates Extreme Sample Filtering as a pre-processing step before this greedy selection, which aims to filter out the samples with extremely perplexing scores that may hinder model learning~\cite{maharana2024adapt, marion2023less}.

Our contributions are summarized as follows:
\begin{itemize}
    \setlength{\itemsep}{0pt}
    \setlength{\parsep}{0pt}
    \setlength{\parskip}{0pt}
\item 
We present a novel framework to mitigate the significant computational overhead for capability gains probing of large-scale ChartU training sets, which leverages subset training to evaluate the enhancement attainable via full-set training.
It helps researchers in making decisions about data generation and  expedites the iteration and updating of datasets.

\item 
We propose a subset selection method, \EXaMCaP, which leverages Entropy Gain Maximization as the criterion to select an informative subset.
It iteratively selects samples with the maximum entropy gain relative to the already selected data, thereby enhancing the overall diversity of the subset to preserve the knowledge coverage of the full ChartU training set.

\item 
We conduct extensive experiments and show that, compared to other competitive methods, \EXaMCaP achieves superior performance in probing the capability gain of the ChartU training set, along with its strong effectiveness across diverse subset sizes and compatibility with different MLLM architectures.

\end{itemize}

\begin{figure*}[htp]
    \centering
    \includegraphics[width=1\textwidth]{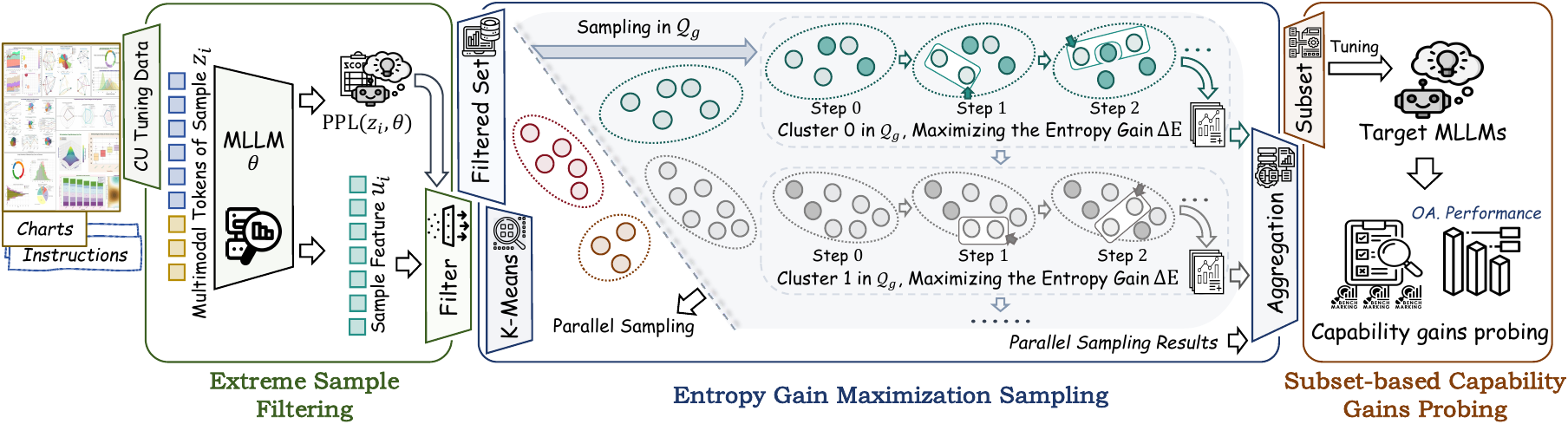}
    \caption{
    The architecture of \EXaMCaP.
    For the given ChartU training set, \EXaMCaP first conducts preliminary extreme sample filtering based on PPL.
    Secondly, it performs entropy gain maximization sampling within each cluster from prior partitioning, and then obtains a subset with high entropy to maintain ChartU knowledge diversity.
    Finally, it performs subset-based capability gains probing.
    }
    \label{fig:method}
\end{figure*}

\section{Related Work}

Chart Understanding (ChartU) requires advanced visual recognition and reasoning capabilities of MLLMs~\cite{yang2025effective, xu2025chartm}. 
Integrating such capabilities enhances MLLMs’ contextual analysis and reasoning abilities for chart information, expanding their applicability in broader domains such as data analytics~\cite{liu2024mmc}.

\paragraph{ChartU Dataset Generation}
Given that enhancing MLLMs’ understanding abilities largely depends on the knowledge provided by the training data~\cite{chen2024far}, recent works~\cite{han2023chartllama, meng2024chartassisstant, zhang2024tinychart} 
focus on constructing automated ChartU datasets followed by visual instruction tuning, including collecting existing charts and then synthesizing instructions~\cite{liu2024mmc, masry2025chartgemma} or jointly synthesizing charts and instructions~\cite{yang2025effective, xu2025chartm, he2024distill}.
After a certain scale of generation, the ChartU training set is fully utilized for fine-tuning and evaluation to assess the capability gains brought by the generated ChartU dataset, which quantifies the enhancement of the ChartU model’s performance provided by the synthetic dataset’s knowledge.
Researchers then decide whether to scale up the generation or directly adopt the current-scale generated data as the final version.

\paragraph{Data Selection for Visual Instruction Tuning}
Recent data selection methods~\cite{wei2023instructiongpt, chen2024your}, 
which are used for selecting a subset from the entire general visual instruction tuning dataset, 
prove the feasibility of training on subsets instead of full datasets.
\citet{chen2024your, yan2025coido} accomplish subset selection by training a score-net.
COINCIDE~\cite{lee2024concept} proposes a clustering method to identify concept-skill compositions.
DataTailor~\cite{yu2025mastering} selects data based on informativeness, uniqueness, and representativeness to retain the most relevant samples.
Prism~\cite{bi2025prism} considers the anisotropy in visual embeddings, corrects them through implicit re-centering for image selection, and identifies the related instructions.
PreSeL~\cite{safaei2025filter} integrates task-importance estimation and task-wise cluster-based selection to obtain the most representative images.
However, direct application of most of these recent methods to ChartU training set capability gain probing remains challenging.
On the one hand, ~\citet{wu2024icons, yu2025mastering, safaei2025filter} require explicit fine-grained task partitioning reflecting different domains to support subsequent selection operations. 
However, recent ChartU training sets~\cite{yang2025effective, xu2025chartm} generate data directly around diverse fine-grained reasoning queries, with only ambiguous partitioning (e.g., description and reasoning) or even no clear partitioning at all.
Furthermore, ~\citet{safaei2025filter, bi2025prism} take images as the entry point for subset selection.
Yet, a single chart in the ChartU training set~\cite{he2024distill, xu2025chartm} is generally linked to multiple instructions, and the overall number of different charts is relatively small.
This leads such methods to discard a large number of equally information-rich charts and reduce the overall coverage of charts.
On the other hand, ~\citet{liu2025less} and ~\citet{wu2024icons} perform subset selection based on gradients, but the iterative computation of these gradients incurs costs comparable to full-set training itself~\cite{bi2025prism}, which runs counter to the original motivation of capability gain probing.

\section{\EXaMCaP}

To identify the most informative subset for probing the capability gains of the ChartU training set, we propose \EXaMCaP.
As shown in Figure~\ref {fig:method}, it contains
three steps: 
Extreme Sample Filtering,
Entropy Gain Maximization Sampling, and Subset-based Capability Gains Probing.

\subsection{Extreme Sample Filtering}
Perplexity (PPL) denotes the average negative log-likelihood of next-label prediction. 
Extreme PPL values (too high or too low) indicate the model’s extreme output token probabilities, which reflects sample difficulty to some extent~\cite{marion2023less, li2024quantity}.
Some studies~\cite{maharana2024adapt, marion2023less} suggest that samples with extreme PPL values (called extreme samples) may impair model training performance. 
Therefore, following previous work~\cite{maharana2024adapt}, \EXaMCaP leverages PPL for preliminary filtering of extreme samples.

For a ChartU training set $\mathcal{D}^0$ where each sample $z_i$ comprises a chart $I_i$ and a textual instruction with answer $a_i$, EXaMCaP inputs $z_i$ to an MLLM $\theta$ 
to obtain the PPL of $a_i$:
\begin{equation}
\text{PPL}(a_i; \theta) = \exp\left( \frac{1}{|a_i|} \sum_{e_j \in a_i} \text{NLL}(e_j) \right),
\end{equation}
where $\text{NLL}(e_j)$ denotes the negative log-likelihood of the token $e_j$ in $a_i$.
\EXaMCaP filters out samples at both ends of the PPL distribution, 
yielding $\mathcal{D} \subseteq \mathcal{D}^0$ with size $K$.

\subsection{Entropy Gain Maximization Sampling}

More diverse training data is vital to boosting MLLMs’ performance~\cite{he2024distill, lee2024concept}, as it provides broader and richer knowledge encompassing more diverse samples and a wider range of reasoning patterns.
Additionally, entropy serves as a metric for quantifying system uncertainty and inherently characterizes distribution diversity~\cite{leinster2020entropy, yang2025effective}.
Motivated by this, \EXaMCaP aims to maximize the entropy of the selected subset to enhance diversity, thereby improving overall knowledge coverage.
Given the infeasibility of enumerating all possible subsets of a predefined size, 
\EXaMCaP iterates over all samples $\boldsymbol{z}_i$ in the candidate sampling space $\mathcal{X}$ at step $t$, denoted as $\boldsymbol{z}_i^t$, to select the sample $\boldsymbol{z}^\ast$ with the maximum entropy gain:
\begin{align}
\boldsymbol{z}^\ast 
&= \underset{\boldsymbol{z}_i^t \in \mathcal{X} \setminus \mathcal{G}_{t,\mathcal{X}}}{\arg\max} \ 
 \Delta\text{E}(\mathcal{G}_{t,\mathcal{X}} \cup \{\boldsymbol{z}_i^t\}) \notag \\
&= \underset{\boldsymbol{z}_i^t \in \mathcal{X} \setminus \mathcal{G}_{t,\mathcal{X}}}{\arg\max} \ 
 \left[ \text{E}(\mathcal{G}_{t,\mathcal{X}} \cup \{\boldsymbol{z}_i^t\}) - \text{E}(\mathcal{G}_{t,\mathcal{X}}) \right],
\label{eq:general_entropy_gain}
\end{align}
where $\mathcal{G}_{t,\mathcal{X}}$ denotes the set of samples already selected from the 
$\mathcal{X}$ at step $t$,
and $\text{E}(\cdot)$ denotes the entropy calculation function for the sample set.

In particular, \EXaMCaP first conducts clustering, then performs greedy sampling within each cluster. 
Additionally, it incorporates computational and memory optimizations to further improve sampling speed.

\paragraph{K-Means Clustering}
Iterative sampling from a large and high-dimensional candidate set $\mathcal{X} = \mathcal{D}$ 
causes prohibitive computational and time costs. 
This inefficiency stems from the necessity of traversing $\mathcal{D} \setminus \mathcal{G}_{t,\mathcal{D}}$ in each iteration to identify the sample that yields the maximum entropy gain.

Clustering aggregates similar samples, with each centroid acting as its cluster’s representative to approximate the full set’s key features. 
To reduce the sampling space and ensure the coverage of the entire set by the sampling space,
\EXaMCaP first conducts K-Means clustering, partitions $\mathcal{S}$ into $L$ clusters $C=\{C_0, C_1, \dots, C_{L-1}\}$, and then performs entropy gain maximization sampling within each cluster, where the candidate space is $\mathcal{X} = C_l$ (the $l$-th cluster).

\paragraph{Eigenvalues Entropy of Similarity Matrix} 
Directly computing the entropy of $\mathcal{G}_{t,\mathcal{X}}$ (e.g., Shannon entropy) is challenging,  as its underlying probability distribution is often unknown a priori.
Von Neumann entropy~\cite{bengtsson2017geometry} offers a viable solution, which quantifies the uncertainty of a quantum system:
\begin{align}
\text{E}(\rho) &= -\operatorname{Tr}\left(\rho \ln \rho\right), \label{eq:von_neumann_entropy} 
\end{align}
where the trace operator $\operatorname{Tr}(\cdot)$ denotes the sum of the eigenvalues of the density matrix $\rho$, which satisfies the properties of positive semi-definiteness and unit trace. 
Following \citet{friedman2022vendi}, we leverage the similarity matrix of $\mathcal{G}_{t,\mathcal{X}}$, denoted as $\mathcal{M}_{\mathcal{G}_{t, \mathcal{X}}}$, to compute this entropy. 
Specifically, considering that $\mathcal{M}_{\mathcal{G}_{t, \mathcal{X}}}$ quantifies inter-sample correlations, 
and the positive definiteness of the Gaussian similarity measure $\exp\left(-\|\boldsymbol{u}_i - \boldsymbol{u}_j\|_2^2 / 2\sigma^2\right)$, the constructed $\mathcal{M}_{\mathcal{G}_{t, \mathcal{X}}}$ is positive semi-definite, 
where $\forall i,j \in \{1,2,\dots,|\mathcal{G}_{t, \mathcal{X}}|\}$ and $\boldsymbol{u}_i$ denotes the average-pooled last-layer embedding of $\boldsymbol{z}_i$ extracted from the specified MLLM.
By performing trace normalization on $\mathcal{M}_{\mathcal{G}_{t,\mathcal{X}}}$, the resulting matrix retains positive semi-definiteness and satisfies the unit trace requirement, fully conforming to the definition of a density matrix:
\begin{align}
\rho_{\mathcal{G}_{t,\mathcal{X}}} &= \frac{\mathcal{M}_{\mathcal{G}_{t,\mathcal{X}}}}{\operatorname{Tr}\left(\mathcal{M}_{\mathcal{G}_{t,\mathcal{X}}}\right)}. \label{eq:trace_norm}
\end{align}

Based on this, 
the entropy of $\mathcal{G}_{t,\mathcal{X}}$ can be computed by substituting Eq.~\ref{eq:trace_norm} into Eq.~\ref{eq:von_neumann_entropy}.
It holds that when the entropy is maximized, the density matrix of the normalized similarity matrix approaches a diagonal matrix and its eigenvalues follow a uniform distribution. 
This implies that all samples in $\mathcal{G}_{t,\mathcal{X}}$ are fully dissimilar, which corresponds to the highest level of diversity. 

\paragraph{Intra-Cluster Greedy Sampling}
Within each cluster $C_l$, to compute the initial entropy, \EXaMCaP first randomly selects two samples to calculate the pairwise similarity, thereby forming the initial set 
$\mathcal{G}_{0,C_l}$. 
It then conducts incremental greedy sampling over the candidate space 
$\mathcal{X} = C_l$
to add the optimal new sample $\boldsymbol{z}^\ast$:
\begin{equation}
\label{eq:add_new_sample}
\begin{split}
\boldsymbol{z}^\ast 
&= \underset{
    \boldsymbol{z}_i^t \in \mathcal{X} \setminus \mathcal{G}_{t, \mathcal{X}}
}{\arg\max} \ \Delta \text{E}\left(\mathcal{G}_{t,\mathcal{X}} \cup \{ \boldsymbol{z}_i^t \}\right) \\
&= \underset{
    \boldsymbol{z}_i^t \in \mathcal{X} \setminus \mathcal{G}_{t, \mathcal{X}}
}
{\arg\max} \Bigg[ 
    \text{E}\bigg(
        \frac{\mathcal{M}_{\mathcal{G}_{t,\mathcal{X}} \cup \{ \boldsymbol{z}_i^t \}}}{\operatorname{Tr}\big(\mathcal{M}_{\mathcal{G}_{t,\mathcal{X}} \cup \{ \boldsymbol{z}_i^t \}}\big)}
    \bigg) \\
    &\quad - \text{E}\bigg(\frac{\mathcal{M}_{\mathcal{G}_{t,\mathcal{X}}}}{\operatorname{Tr}(\mathcal{M}_{\mathcal{G}_{t,\mathcal{X}}})}\bigg) \Bigg].
\end{split}
\end{equation}

To efficiently update the $\mathcal{M}_{\mathcal{G}_{t,\mathcal{X}} \cup \{ \boldsymbol{z}_i^t \}}$, \EXaMCaP incrementally appends rows and columns corresponding to the embedding $\boldsymbol{u}_i^t$ of the candidate sample $\boldsymbol{z}_i^t$ to the existing matrix $\mathcal{M}_{\mathcal{G}_{t,\mathcal{X}}}$:
\begin{equation}
\mathcal{M}_{\mathcal{G}_{t,\mathcal{X}} 
\cup 
\{ \boldsymbol{z}_i^t \}}
= 
\begin{bmatrix} 
    \mathcal{M}_{\mathcal{G}_{{t,\mathcal{X}}}} & \boldsymbol{s}_i^t \\ (\boldsymbol{s}_i^t)^\top & 1 
\end{bmatrix},
\label{eq:incr_matrix_update}
\end{equation}
where $\boldsymbol{s}_i^t$ denotes the similarity vector, with each entry representing the Gaussian similarity between the embedding $\boldsymbol{u}_i^t$ and the embeddings $\boldsymbol{u}_j$ of the selected samples in $\mathcal{G}_{t,\mathcal{X}}$.

For the budget of the $C_l$, \EXaMCaP allocates cluster-wise budgets proportional to cluster size:
\begin{equation}
B_l = \max\left(1, \left\lfloor \frac{|C_l|}{\sum_{l=0}^{L-1} |C_l|} \cdot B \right\rfloor\right) ,
\label{eq:cluster_budget}
\end{equation}
where $B_l$ denotes the sampling budget for cluster $C_l$, and $B$ is the total sampling budget. 
This allocation avoids introducing additional hyperparameters.
Additionally, the entropy gain maximization of the diversity-preserving sampling scheme in Eq.~\ref{eq:general_entropy_gain} is highly suitable for scenarios where larger clusters contain prevalent, hard-to-partition data with abundant core features.
Additionally, we adopt a relatively large $L$ to ensure centroids cover the dataset’s global core features and mitigate the potentially implicit excessive inter-cluster size imbalance induced by proportional allocation.

\paragraph{Computational and Memory Optimizations}

We integrate parallel computation and memory optimization to further boost sampling speed. 
Specifically, \EXaMCaP first partitions the cluster set $\{C_0, C_1, \dots, C_{L-1}\}$ into $G$ disjoint subsets $\{\mathcal{Q}_0, \mathcal{Q}_1, \dots, \mathcal{Q}_{G-1}\}$ and assigns each subset $\mathcal{Q}_g$ to GPU $g$ for independent intra-cluster sampling, thus enabling parallelized sampling across $G$ available GPUs. 
Finally, we further reduce the candidate set size for intra-cluster greedy sampling by computing the entropy gain only on a preset random candidate set $\mathcal{X}=r_t$ of fixed size $m$ at each traversal step $t$, where where $r_t \subseteq C_l \setminus \mathcal{G}_{t,C_l}$, $|r_t|=m$. 
For the algorithm of complete Entropy Gain Maximization Sampling, please refer to Appendix~\ref{appendix:alg}.

\subsection{Subset-based Capability Gains Probing}
Once the subset $\mathcal{G}_{final}$ is obtained, \EXaMCaP conducts subset substitutive probing by training target MLLMs on $\mathcal{G}_{final}$ instead of the full ChartU training set.
Furthermore, the selected $\mathcal{G}_{\text{final}}$ is compatible with MLLM across various architectures, enabling fine-tuning to probe performance improvements in CharU on the full dataset.
Our ablation experiments validate this feasibility.
The complete \EXaMCaP workflow helps boost capability gain probing efficiency for the ChartU training set, aiding researchers decide whether to scale to the next data size or explore new data generation directions, and thus facilitating iterative refinement of the ChartU training set.

\begin{table*}[]
\caption{
Main experimental results for capability gains probing of ChartU training sets. 
\textsc{Out-of-Domain} denotes benchmarks independent of the corresponding training set, while \textsc{In-Domain} denotes benchmarks native to the training set. 
The symbol ``$^\dag$'' indicates that ``\textsc{Avg-Rel}'' is the average of relative performance across all benchmarks, 
, across benchmarks to assess the level of generalization.
}
\centering
\resizebox{\textwidth}{!}{
\begin{tabular}{clcccccccccccc}
\toprule
& \multicolumn{1}{c}{} 
& \multicolumn{7}{c}{\textsc{\textsc{Out-of-Domain}}} 
&  
& \multicolumn{3}{c}{\textsc{\textsc{In-Domain}}} 
&  \\ 
\cmidrule(lr){3-9} \cmidrule(lr){11-13}

& \multicolumn{1}{c}{} 
& \multicolumn{2}{c}{CharXiv} 
& \multicolumn{2}{c}{ChartQA} 
& ChartX 
& \multicolumn{2}{c}{ReachQA} 
&  
& \multicolumn{2}{c}{$\text{ECD}\,_\text{Bench}$} 
& $\text{MMC}\,_\text{Bench}$ 
&  \\

\cmidrule(lr){3-4}
\cmidrule(lr){5-6}
\cmidrule(lr){7-7}
\cmidrule(lr){8-9}
\cmidrule(lr){11-12}
\cmidrule(lr){13-13}

\multirow{-3}{*}{
    \begin{tabular}[c]{@{}c@{}}\textsc{Train Set}\\ \textsc{(Size)}\end{tabular}
} 
& \multicolumn{1}{c}{
    \multirow{-3}{*}{\textsc{Methods}}
} 
& reas. & desc. & human & augmented & reas. & reas. & reco. 
& \multirow{-3}{*}{
    \textsc{\begin{tabular}[c]{@{}c@{}}Avg-Rel\\ (\%)\end{tabular}}
} 
& reas. & desc. & MQA 
& \multirow{-3}{*}{
    \textsc{\begin{tabular}[c]{@{}c@{}}Avg-Rel$^\dag$ \\ (\%)\end{tabular}}
} \\ 

\midrule

\rowcolor[HTML]{E7E6E6} 
- & Original & 17.80 & 37.92 & 54.72 & 78.48 & 28.73 & 7.30 & 21.80 & - & 8.17 & 18.95 & 71.16 & - \\ 

\midrule

\rowcolor[HTML]{F2F2F2} 
\begin{tabular}[c]{@{}c@{}}MMC$_{\text{Ins.}}$\\ (410K)\end{tabular} 
& Full Set & 21.70 & 33.55 & 45.60 & 73.36 & 28.91 & 7.20 & 23.50 & 100.00 & - & - & 70.54 & 100.00 \\

& Random & 19.30 & 32.99 & \textbf{47.28} & 77.20 & {\ul 28.65} & 6.10 & 20.50 & 95.32 & - & - & 72.52 & 96.26 \\

& Perplexity & 19.60 & 31.91 & 42.96 & 75.12 & 27.43 & 6.30 & 19.00 & 92.18 & - & - & {\ul 75.74} & 94.08 \\

& EL2N & 19.90 & \textbf{36.29} & 44.48 & 63.60 & 24.39 & \textbf{6.90} & 19.40 & 92.41 & - & - & 64.60 & 92.31 \\

& CCS & \textbf{20.90} & 33.14 & 45.68 & {\ul 77.28} & 26.91 & {\ul 6.60} & \textbf{21.30} & {\ul 96.57} & - & - & 63.36 & 95.73 \\

& COINCIDE & {\ul 20.00} & 32.29 & 46.16 & 74.96 & 28.56 & 6.50 & {\ul 20.90} & 95.69 & - & - & 68.56 & 95.88 \\

\multirow{-6}{*}{\begin{tabular}[c]{@{}c@{}}MMC$_{\text{Ins.}}$\\ (80K)\end{tabular}} 

& Ours & 19.70 & {\ul 33.43} & {\ul 46.96} & \textbf{79.44} & \textbf{31.25} & \textbf{6.90} & 20.60 & \textbf{99.04} & - & - & \textbf{76.61} & \textbf{100.24} \\ 

\midrule

\rowcolor[HTML]{F2F2F2} 
\begin{tabular}[c]{@{}c@{}}ECD\\ (321K)\end{tabular} 
& Full Set & 22.80 & 57.28 & 61.68 & 78.00 & 50.69 & 15.50 & 40.20 & 100.00 & 23.04 & 51.96 & - & 100.00 \\

& Random & {\ul 22.90} & 54.15 & 59.04 & {\ul 80.88} & {\ul 45.75} & 10.40 & 33.30 & 90.65 & 14.87 & 38.07 & - & 85.82 \\

& Perplexity & 22.70 & 54.32 & 56.56 & 79.04 & 41.49 & 10.00 & 30.20 & 86.99 & 15.11 & \textbf{41.42} & - & 83.80 \\

& EL2N & 21.90 & 51.69 & \textbf{61.20} & {\ul 80.88} & \textbf{46.35} & {\ul 11.90} & \textbf{34.90} & {\ul 92.03} & {\ul 16.18} & {\ul 40.03} & - & {\ul 87.94} \\

& CCS & 22.60 & \textbf{55.27} & 58.08 & 74.88 & 43.58 & 10.50 & 31.40 & 88.23 & 15.20 & 39.79 & - & 84.46 \\

& COINCIDE & 22.30 & {\ul 54.67} & 58.24 & 79.44 & 43.92 & 11.10 & 31.70 & 89.52 & 15.60 & 39.54 & - & 85.60 \\

\multirow{-6}{*}{\begin{tabular}[c]{@{}c@{}}ECD\\ (60K)\end{tabular}} 
 
& Ours & \textbf{23.10} & 54.16 & {\ul 60.00} & \textbf{81.84} & 44.79 & \textbf{12.00} & {\ul 34.80} & \textbf{92.92} & \textbf{18.14} & {\ul 40.03} & - & \textbf{89.58} \\
\bottomrule
\end{tabular}
}

\label{tab:mainResults}
\end{table*}

\section{Experimental Setup}

\paragraph{Datasets and Metrics}
We conduct capability gains probing on two distinct instruction training sets for ChartU: \\
MMC$_{\text{Instruction}}$ 
(410K) is mainly built by curating existing charts and synthesizing matched instructions, 
which serves as the dedicated dataset for MMC’s~\cite{liu2024mmc} visual instruction tuning. \\
ECD
~\cite{yang2025effective} (321K) is built by jointly synthesizing charts and instructions.
Renowned for high diversity and quality, it covers 29 distinct chart types, including many with subplots.

We employ a wide range of benchmarks and use GPT-4o~\cite{jaech2024openai} as the evaluation tool to extract answers and verify correctness, following ECD. 
As each benchmark varies in scale, following~\citet{lee2024concept}, we calculate the average relative accuracy (\textsc{Avg-Rel}) across all benchmarks, with values relative to those of models trained on the full set.
Please refer to Appendix~\ref{appendix:exp_datasets_metrics} for more details.

\paragraph{Baselines}

We compare \EXaMCaP with several data selection methods that can select subsets to replace full-set training for capability gains probing. 
We first train a model on the complete dataset, designate it as Full Set, and use it as the reference for \textsc{Avg-Rel} comparisons. 
We also establish the Random baseline by training a model on a randomly selected subset.
In addition, we incorporate three baselines, including two score-based methods Perplexity~\cite{marion2023moreinvestigatingdatapruning} and EL2N~\cite{paul2021deep}, and a coverage-based method CCS~\cite{zheng2023coverage}. 
We also include COINCIDE~\cite{lee2024concept} as a baseline. 
It is a state-of-the-art method developed prior to the methods~\cite{bi2025prism, yu2025mastering, wu2024icons} that are not readily applicable to capability gain probing on ChartU training sets.

\paragraph{Implementation Details}
We use LLaVA-Next-LLaMA3-8B~\cite{li2024llava} as the primary base model
, following the experimental settings commonly adopted in recent ChartU dataset generation works~\cite{he2024distill, yang2025effective}.
We adopt LoRA~\cite{hu2022lora} to train the model for 1 epoch with a batch size of 8. 
Additionally, we set the cluster number $L=1000$ and the random candidate set $m=100$ as the default settings.
Please refer to Appendix~\ref{appendix:exp_baselines} for more details.

\section{Experimental Results}

\subsection{Main Results}

Table~\ref{tab:mainResults} summarizes the capability gain probing results of MMC$_{\text{Ins.}}$ and ECD, where results characterize the performance of the fine-tuned LLaVA-Next-LLaMA3 on several downstream benchmarks.

We notice that, in terms of \textsc{Avg-Rel} on the two ChartU training sets, 
\EXaMCaP consistently demonstrates superior and stable probing performance, achieving $99.04\%$ and $92.92\%$ of the full-set training level, respectively.
Additionally, we observe that \EXaMCaP produces 
average relative accuracy 
that surpass those of full-set training when evaluated on the in-domain benchmark MQA of MMC$_{\text{Bench}}$, combined with the \textsc{Avg-Rel} of out-of-domain.
This phenomenon arises because most instructions in MQA are of a multiple-choice reasoning type, which has a data format that differs significantly from that of MMC$_{\text{Ins.}}$.
An overabundance of tuning data with similar formats can interfere with the MLLM's inherent reasoning capabilities, resulting in reduced performance on multiple-choice questions.

Furthermore, when compared to other competitive methods, \EXaMCaP consistently maintains top-tier performance with the full-set training method across all benchmark tests.
Specifically, it achieves either the best performance or ranks a close second in 6 out of 8 benchmark tests on MMC and 7 out of 9 benchmark tests on ECD.
In contrast, we observe that some methods excel in probing performance on certain benchmarks, yet perform poorly on others, with variations depending on the training set.
We hypothesize that this occurs because score-based methods like EL2N are subject to selection biases induced by inherent model capabilities, which in turn leads to insufficient data coverage. 
For the COINCIDE, it adopts an intra-cluster sampling strategy designed to align with the original intra-cluster distribution, 
which tends to favor densely distributed regions while overlooking sparse yet valuable diverse data within clusters.
The results reveal that \EXaMCaP enables more stable probing of capability gains compared to these competitive methods.

To sum up, rather than relying solely on the full ChartU dataset, a small subset selected by \EXaMCaP, approximately 20\% of the full set, can effectively probe the capability gains of the entire training set.

\subsection{Further Analysis}
\label{sec:further_analysis}

In this section, we perform experiments on ECD to evaluate the effectiveness of \EXaMCaP across different sizes of subset and various structures of MLLMs.
For more detailed presentations of the results, please refer to Appendix~\ref{appendix:exp_futher}.

\paragraph{Analysis of the Subset Size}

\begin{figure}[t]
  \includegraphics[width=\columnwidth]{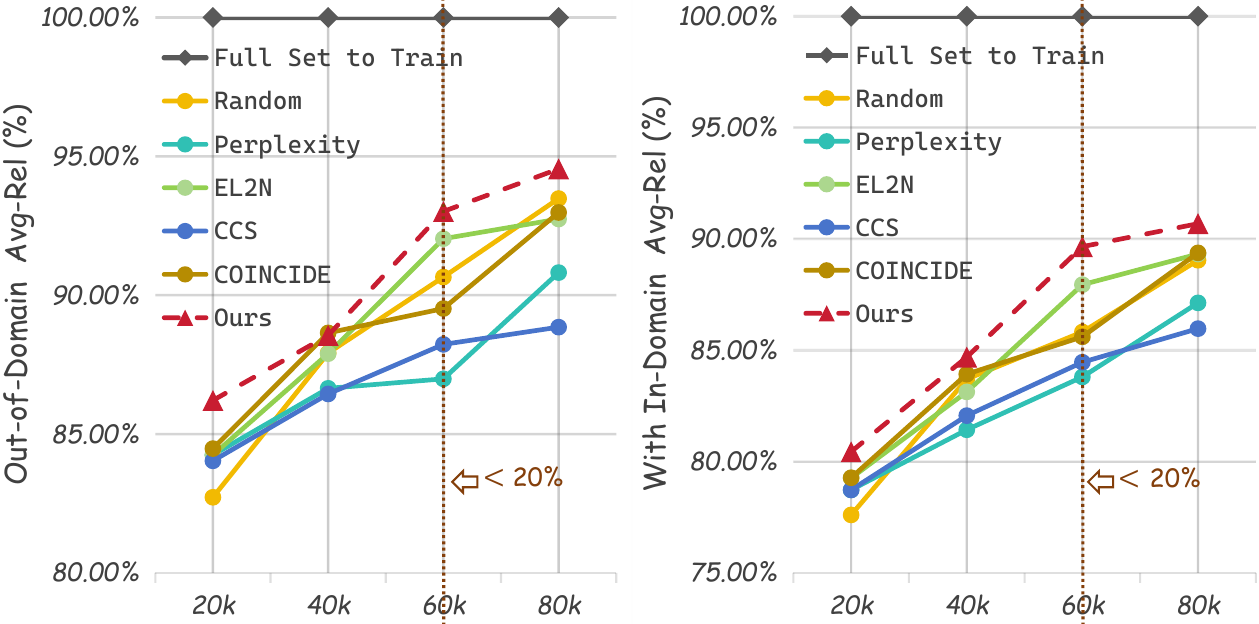}
   \caption{The impact of subset sizes on capability gain probing. The subset sizes range from 20K to 80K with an interval of 20K, and 60K acts as the dividing line for subsets that account for approximately 20\% or less of the total training set size.}
  \label{fig:exp_diff_sizes}
\end{figure}

\begin{table*}[]
\small
\caption{Ablation study on specific design choices. 
The row highlighted with \colorbox[HTML]{D6DCE4}{~~} corresponds to the unaltered \EXaMCaP results. 
The middle pane present ablation results of \EXaMCaP regarding extreme sample filtering and cluster budget allocation strategies.
The last pane demonstrates the impacts of different intra-cluster sampling strategies on the performance of capability enhancement probing.
}

\centering
\begin{tabular}{ccccc}
\toprule
\textsc{
    \begin{tabular}[c]{@{}c@{}}Extreme Sample \\ Filtering\end{tabular}
} 
& \textsc{
    \begin{tabular}[c]{@{}c@{}}Cluster Budget \\ Allocation\end{tabular}
} 
& \textsc{
    \begin{tabular}[c]{@{}c@{}}Intra-Cluster \\ Sampling\end{tabular}
} 
& \textsc{
    \begin{tabular}[c]{@{}c@{}}Out-of-Domain \\ Avg-Rel (\%)\end{tabular}
}
& \textsc{
    \begin{tabular}[c]{@{}c@{}}With In-Domain \\ Avg-Rel (\%)\end{tabular}
} \\

\midrule
\rowcolor[HTML]{F2F2F2} 
\multicolumn{3}{c}{Full Set to Train} & 100.00 & 100.00 \\
\midrule
\rowcolor[HTML]{D6DCE4}
\checkmark
& Proportion
& \textsc{EXaM} & 92.92 & 89.58 \\
\midrule
- & Proportion & & 92.74 & 88.94 \\
\checkmark & Average & \multirow{-2}{*}{\textsc{EXaM}} & 92.52 & 87.82  \\
\midrule

&  & Random & 90.30 & 86.51 \\
\multirow{-2}{*}{\checkmark} 
& \multirow{-2}{*}{Proportion} & MMD$^2$-Minimize & 89.63 & 85.68 \\
\bottomrule
\end{tabular}

\label{tab:ablation_component}
\end{table*}

\begin{figure}[t]
  \includegraphics[width=\columnwidth]{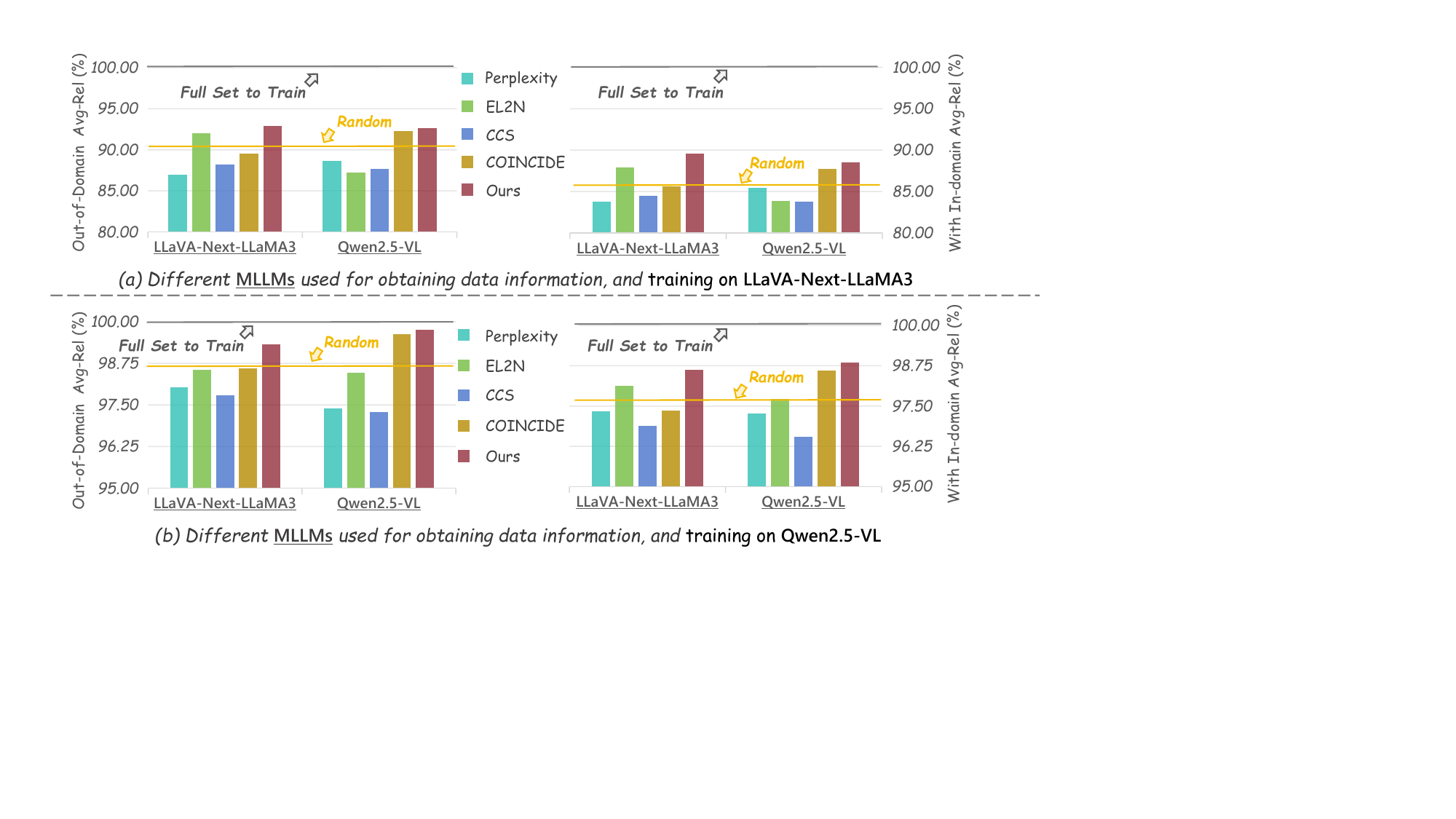}
   \caption{The impact of alternative extraction and target MLLMs on capability gains probing, involving using different MLLMs obtain data information (such as embeddings), then using (a) LLaVA-Next-LLaMA3 and (b) Qwen2.5-VL for fine-tuning.
   }
  \label{fig:exp_diff_mllms}
\end{figure}

We investigate the impact of subset sizes on capability gain probing by using LLaVA-Next-LLaMA3 as the primary base model.
Figure~\ref{fig:exp_diff_sizes} shows the results across different subset sizes.
We find that \EXaMCaP consistently outperforms other methods across various sizes, highlighting the effectiveness of our method.
When the subset size is limited to 20\% (60K) or less of the full set, \EXaMCaP achieves over 80\% for the \textsc{Avg-Rel} including that on in-domain tasks.
Furthermore, with only about 6\% of the full training set data, \EXaMCaP reaches an \textsc{Avg-Rel} exceeding 85\%  on out-of-domain tasks, while other methods show significantly lower probing performance.
For subset sizes greater than 20\% (80K) of the full set, we observe modest improvements in \textsc{Avg-Rel}, yet such sizes incur higher tuning time costs.

\paragraph{Alternative Various Structures of MLLMs}
To investigate the impact of data information extracted from different MLLMs on data selection, as well as the capability gains of different MLLMs brought by the selected subsets, we alternately employ LLaVA-Next-LLaMA3-8B and Qwen2.5-VL-7B as either the source for data information extraction or the target for fine-tuning. 
The training parameters for these models are consistent with those used in the main experiments.
As shown in Figure~\ref{fig:exp_diff_mllms}, \EXaMCaP shows its model robustness since the selected subset, which is derived from data information obtained by different MLLMs, achieves superior \textsc{Avg-Rel} whether it is used for fine-tuning the MLLM itself or other MLLMs.

Additionally, we observe that fine-tuning two MLLMs with diverse architectures and ChartU capabilities on the same training set yields variable \textsc{Avg-Rel} against full-set training, with higher values for Qwen2.5-VL and lower values for LLaVA-Next-LLaMA3. 
This variation can be attributed to differences in the breadth and depth of ChartU knowledge acquired by the MLLMs during pre-training, as evidenced by the vanilla Qwen2.5-VL reported in~\citet{bai2025qwen2} achieving better performance on the ChartQA test set via prompting alone than the vanilla LLaVA-Next-LLaMA3 reported in~\citet{li2024llava}. 
Consequently, there are notable differences in how much ChartU knowledge is supplemented and the performance gains realized from the newly generated training set.
This variable comparability to full-set training, as reflected by \textsc{Avg-Rel}, enables \EXaMCaP to validate more rapidly than full-set training whether the generated training set can supply sufficient ChartU knowledge to MLLMs with diverse architectures and capabilities, including those with inherently strong ChartU capabilities.
In engineering practice, the original performance of the MLLMs to be fine-tuned is easily accessible. The process of selecting a subset using \EXaMCaP, conducting the fine-tuning, and comparing the resulting performance to the original baseline enables researchers to conduct capability probing more efficiently.

\subsection{Ablation}
\label{sec:ablation}

In this section, we conduct ablation studies to isolate the effects of specific design choices and investigate the impact of different hyperparameter settings on the effectiveness of capability gain probing. 
We perform ablation experiments on the ECD training set with LLaVA-Next-LLaMA3. 
For more detailed presentations of the results, please refer to Appendix~\ref{appendix:exp_ablation}. 

\paragraph{Ablation on the Specific Design Choices}
We isolate the effects of specific design choices from three perspectives, including extreme sample filtering, cluster budget allocation methods, and intra-cluster sampling methods.
The results are summarized in Table~\ref{tab:ablation_component}.

For extreme sample filtering, we find that excluding extreme samples can negatively affect the final performance. 
One of the main reasons for this is that samples with extremely low or high perplexity values may correspond to outliers, low-quality samples, and uninformative data \cite{zheng2023coverage, maharana2024adapt}.
We also showcase several cases in Appendix~\ref{appendix:extreme_samples} and find that some samples with excessively high perplexity values indeed suffer from issues such as explicit chart overlaps and annotation errors.

For cluster budget allocation, we compare the Proportion (refer to Eq.~\ref{eq:cluster_budget}) allocation strategy with another commonly used method called Average. We follow the Average described in \citet{maharana2024adapt}, which does not require any additional hyperparameter tuning to adapt to various training sets and models.
We observe that direct Average allocation is slightly inferior to Proportion allocation based on cluster size.
We argue that this phenomenon is influenced by the characteristics of the ChartU benchmark, as it may contain a higher prevalence of common chart types. 
The Proportion allocation method involves more sampling in larger clusters that contain these common charts and reasoning queries. 
This approach helps the model learn this common knowledge more efficiently, ultimately improving its performance in downstream benchmarks.

\begin{table}[]
\caption{The impact of cluster numbers $L$ and random candidate set size $m$ on capability gains probing.}
\centering
\small
\begin{tabular}{lcc}
\toprule

\textsc{Methods} & \textsc{\begin{tabular}[c]{@{}c@{}}Out-of-Domain\\ Avg-Rel (\%)\end{tabular}} & \textsc{\begin{tabular}[c]{@{}c@{}}With In-Domain\\ Avg-Rel (\%)\end{tabular}} \\ \midrule
\rowcolor[HTML]{F2F2F2} 
Full Set & 100.00 & 100.00 \\ 
\midrule
\rowcolor[HTML]{E4EAEA} \multicolumn{3}{l}{{\ul {\textit{Fix $m=100$ , change $L$}} }}\\
$L=500$ & 92.41 & 88.41 \\
$L=1000$ & 92.92 & 89.58 \\
$L=1500$ & 92.37 & 89.19 \\ 
\midrule
\rowcolor[HTML]{E4EAEA} \multicolumn{3}{l}{{\ul {\textit{Fix $L=1000$ , change $m$}} }}\\
$m=10$ & 91.82 & 87.83 \\
$m=100$ & 92.92 & 89.58 \\
$m=1000$ & 93.08 & 89.48 \\

\bottomrule
\end{tabular}

\label{tab:ablation_hyper}
\end{table}

\begin{figure}[t]
  \includegraphics[width=\columnwidth]{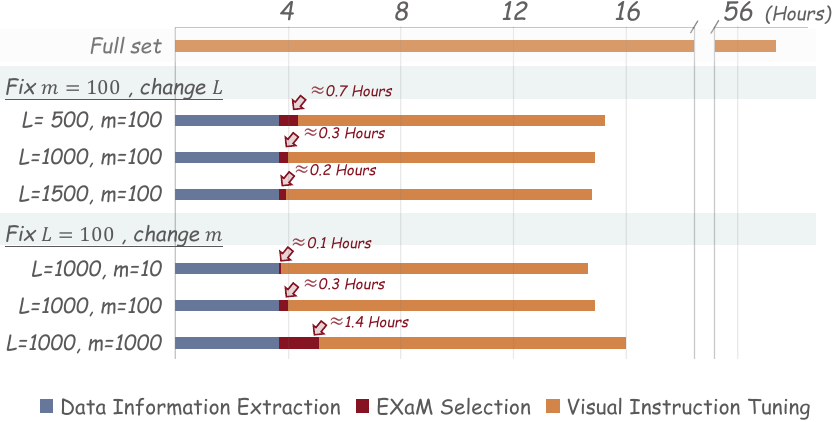}
   \caption{Comparison of time costs under different configurations. 
   The time cost is measured in hours on a computing node with 4 × RTX 4090 GPUs.
    }
  \label{fig:abla_hyper_time_cost}
\end{figure}

For intra-cluster sampling methods, we examine the effects of two alternative techniques, Random selection and MMD$^2$-Minimize~\cite{kim2016examples, lee2024concept}.
As shown in the last panel of Table~\ref{tab:ablation_component}, \EXaM demonstrates superior performance in capability gain probing, while Random yields suboptimal results, and MMD$^2$-Minimize performs slightly worse.
We attribute this outcome to the fundamental goal of MMD$^2$-Minimize, which is to align with the original intra-cluster distribution, yet tends to favor densely distributed regions while overlooking sparse yet valuable and diverse data within clusters.
In contrast, \EXaM samples data based on the principle of maximizing the overall entropy gain of the selected subset, a design that enables the selection of a more diverse and representative subset.
Additionally, \EXaMCaP enhances \EXaM by incorporating extreme sample filtering, which removes unique but potentially low-quality samples. This combination further improves the effectiveness of \EXaM and leads to the best capability gain probing performance on the ChartU training set.
We analyze the entropy gain within clusters for various methods, along with a visualization of the distribution of selected samples. 
Due to space constraints, please refer to the Appendix~\ref{appendix:exp_analysis_vis} for this detailed results and analyses.

\paragraph{Ablation on the Hyperparameter Settings}

We conduct ablation studies on the hyperparameters of \EXaMCaP, including the cluster number $L$ and the random candidate set size $m$. 
As shown in Table~\ref{tab:ablation_hyper}, 
we find that choosing a sufficiently large value (e.g., 1000 or 1500) for $L$ leads to a moderate improvement in the probing performance of \EXaMCaP. 
In contrast, using a relatively smaller $L$ partially degrades the performance, as indicated by the comparison between $88.41\%$ at L=500 and $89.65\%$ at L=1000. 
This can be attributed to the fact that a larger number of clusters allows for finer partitioning of ChartU knowledge, endowing each cluster with higher representativeness and more similar data samples within it.
Sampling conducted with \EXaM thus ensures coverage of globally representative samples while selecting more diverse subsets from similar data samples.
Additionally, we find that an excessively small $m$ leads to biased subset selection, whereas a moderate increase in $m$ can significantly mitigate such bias, thereby boosting the performance of capability gain probing.
Furthermore, we provide a comparison of time costs under different configurations, as shown in Figure~\ref{fig:abla_hyper_time_cost}.
It can be seen that a moderate $L$ and $m$ enable the maintenance of excellent capability gain probing performance while achieving acceptable time costs.

\section{Conclusion}

To mitigate the substantial computational overhead of probing capability gains by fine-tuning MLLMs on the full large-scale ChartU dataset, we propose \EXaMCaP, which identifies the most informative subset to substitute for the full set in the probing process. 
\EXaMCaP builds upon the principle of \uline{E}ntropy \uline{Ga}in \uline{M}aximization to select a high entropy subset, thereby preserving the coverage of ChartU’s knowledge in the full set.
Extensive experiments show that, compared to other competitive methods, \EXaMCaP achieves superior performance in probing the capability gains of ChartU’s training set, while demonstrating strong effectiveness across diverse subset sizes and compatibility with different MLLM architectures.
These results validate \EXaMCaP’s efficacy in probing the capability gain of ChartU’s training process, helping researchers inform decisions on chart dataset generation and accelerate dataset iteration and updates.

\nocite{langley00}

\bibliography{example_paper}
\bibliographystyle{icml2026}

\newpage
\appendix
\onecolumn
\section{Algorithm of Complete Entropy Gain Maximization Sampling (\EXaM)}
\label{appendix:alg}

\begin{algorithm}[tp]
\caption{\EXaM}
\label{alg:alg_exam}
\small
\begin{algorithmic}[1]
    \REQUIRE 
        Filtered set $\mathcal{D}$ with average-pooled embeddings $\{\boldsymbol{u}_i\}_{i=1}^K$, 
        number of clusters $L$, 
        total sampling budget $B$, 
        random candidate set size $m$,
        available GPUs number $G$
    \ENSURE 
        Sampled subset $\mathcal{G}_{\text{final}}$

    \STATE \textcolor[HTML]{6A9955}{/*K-Means Clustering and Budget Allocation*/}
    \STATE $\{C_0, C_1, \dots, C_{L-1}\} \gets \text{Cluster}(\mathcal{D}, L)$
    \STATE $B_{l} \gets$ budget allocation via Eq.~\ref{eq:cluster_budget}

    \STATE \textcolor[HTML]{6A9955}{/*Parallel Cluster Assignment*/}
    \STATE $\{\mathcal{Q}_0, \dots, \mathcal{Q}_{G-1}\} \gets \text{Partition}\left(\{C_0, \dots, C_{L-1}\}, G\right)$
    \STATE $\text{Assign each } \mathcal{Q}_g \text{ to GPU } g \text{ for parallel sampling}$

    \STATE \textcolor[HTML]{6A9955}{/*Intra-Cluster Greedy Entropy Gain Sampling*/}
    \FOR{each GPU $g \in \{0,1,\dots,G-1\}$ \textbf{in parallel}}
        \FOR{each cluster $C_l \in \mathcal{Q}_g$}
            \STATE $\mathcal{G}_{0,C_l} \gets \text{Random}\left(C_l,2\right)$
            \STATE $\text{E}(\mathcal{G}_{0,C_l}) \gets$ compute entropy via Eqs.~\ref{eq:von_neumann_entropy}--\ref{eq:trace_norm}
            \FOR{$t = 1$ to $B_l - 2$}
                \STATE $r_t \gets \text{Random}\left(C_l \setminus \mathcal{G}_{t,C_l}, m\right)$
                \STATE $\Delta\text{E}_{\boldsymbol{z}_i^t} \gets$ entropy gain for each $\boldsymbol{z}_i^t \in r_t$ (computed via Eqs.~\ref{eq:add_new_sample}--\ref{eq:incr_matrix_update})
                \STATE $\boldsymbol{z}^\ast \gets \underset{\boldsymbol{z}_i^t \in r_t}{\arg\max} \Delta\text{E}_{\boldsymbol{z}_i^t}$

                \STATE $\mathcal{M}_{\mathcal{G}_{t,C_l}} = \mathcal{M}_{\mathcal{G}_{t,C_l} \cup \{z^*\}}$
                \STATE $\mathcal{G}_{t,C_l} = \mathcal{G}_{t,C_l} \cup \{z^*\}$

            \ENDFOR

        \ENDFOR

        \STATE $\text{Collect all } \mathcal{G}_{B_l-2,C_l} \text{ for clusters in } \mathcal{Q}_g \text{ as } \mathcal{G}_g$
    \ENDFOR
    
    \STATE $\mathcal{G}_{\text{final}} = \bigcup_{g=0}^{G-1} \mathcal{G}_g$
    \STATE \textbf{Return}  $\mathcal{G}_{\text{final}}$
\end{algorithmic}
\end{algorithm}
We show the complete Entropy Gain Maximization (\EXaM) sampling algorithm in Algorithm \ref{alg:alg_exam}, which integrates intra-cluster greedy sampling for incremental selection of maximum entropy gain samples and accompanying computational and memory optimizations.

\section{Details of Experimental Setups}
\label{appendix:exp_setting_details}
\subsection{Datasets and Metrics}
\label{appendix:exp_datasets_metrics}

\paragraph{ChartU Datasets for Visual Instruction Tuning}
We conduct capability gain probing on two distinct instruction training sets for ChartU:
\begin{itemize}
    \setlength{\itemsep}{0pt}
    \setlength{\parsep}{0pt}
    \setlength{\parskip}{0pt}
\item 
\textbf{MMC$_{\text{Instruction}}$} (410K) is constructed primarily by curating existing charts and synthesizing matched instructions, and it serves as the publicly available portion of the dedicated dataset for MMC’s~\cite{liu2024mmc} visual instruction tuning. 
As one of the early-developed ChartU training datasets built by curating existing charts and synthesizing corresponding instructions, MMC has gained wide adoption and recognition, which is exemplified by its visual instruction tuning subset being used for pre-training by InternVL-1.5~\cite{chen2024far}.
 
\item 
\textbf{ECD}~\cite{yang2025effective} (321K) is a recently proposed ChartU training set for visual instruction tuning, constructed by jointly synthesizing charts and instructions. 
Renowned for its high diversity and quality, it covers 29 distinct chart types, including many with subplots. 
Notably, the original ECD paper states that using the full ECD training set for visual instruction tuning across diverse MLLMs significantly and consistently improves the performance of MLLMs with various architectures on multiple ChartU benchmarks, and compared to most well-known ChartU training sets, ECD-based visual instruction tuning delivers superior performance gains on LLaVA-Next-LLaMA3.
\end{itemize}

\paragraph{ChartU Benchmarks}
We employ a diverse set of ChartU benchmarks to more extensively probe the ChartU capability gains of MLLMs trained on subsets.
We test the trained MLLMs across six benchmarks: two real-world datasets (CharXiv~\cite{wang2024charxiv} and ChartQA~\cite{masry2022chartqa}) and four synthetic datasets (ChartX~\cite{xia2025chartx}, ReachQA~\cite{he2024distill}, ECD$_\text{Bench}$~\cite{yang2025effective}, and MMC$_\text{Bench}$~\cite{liu2024mmc}).

\begin{itemize}
    \setlength{\itemsep}{0pt}
    \setlength{\parsep}{0pt}
    \setlength{\parskip}{0pt}

\item 
\textbf{CharXiv}~\cite{wang2024charxiv} comprises 2,323 challenging charts extracted from arXiv papers, paired with 4,000 descriptive questions about basic chart elements and 1,000 reasoning questions requiring complex visual analysis.

\item 
\textbf{ChartQA}~\cite{masry2022chartqa} is split into an augmented subset (synthetically generated) and a human subset (human-authored queries with greater real-world complexity and advanced reasoning demands). Its test set contains 1,509 chart images paired with 2,500 questions.

\item 
\textbf{ChartX}~\cite{xia2025chartx} contains 1,152 test chart images, each paired with a single question-answer (QA) pair. These charts span 22 disciplinary topics and 18 distinct chart types.

\item 
\textbf{ReachQA}~\cite{he2024distill} consists of 500 high-quality chart images, each paired with two QA pairs—one focusing on basic recognition and the other requiring complex reasoning—totaling 2,000 QA pairs.

\item 
\textbf{MMC$_\text{Bench}$}~\cite{liu2024mmc} is the dedicated evaluation benchmark paired with MMC$_{\text{Instruction}}$. 
Its publicly available portion contains 2,126 true/false binary judgment questions and 808 multiple-choice questions for model benchmarking. We select its multiple-choice question subset, denoted as MQA, as our evaluation dataset. Notably, although MMC$_\text{Bench}$ is affiliated with MMC$_{\text{Instruction}}$, its question-answer format (multiple choice) is inconsistent with that of the training data in MMC$_{\text{Instruction}}$, which adopts a direct question-answer format.

\item 
\textbf{ECD$_\text{Bench}$}~\cite{yang2025effective} is the dedicated evaluation benchmark for ECD. 
It contains 1,224 high-quality chart images, each paired with two QA pairs—one descriptive and one involving complex reasoning—totaling 2,448 QA pairs.

\end{itemize}

As each evaluation benchmark varies in scale, we calculate the average relative accuracy (\textsc{Avg-Rel}) across all benchmarks, following~\cite{lee2024concept}.
\textsc{Avg-Rel} is derived from the formula: 
\begin{equation}
\text{Avg-Rel} = \frac{100\%}{|\mathcal{B}|} \sum_{i \in \mathcal{B}} \frac{\text{Benchmark}^i_{\text{MLLM}_{\text{subset}}}}{\text{Benchmark}^i_{\text{MLLM}_{\text{full-set}}}},
\label{eq:avg_rel}
\end{equation}
where $\mathcal{B}$ denotes the set of all ChartU evaluation benchmarks,
$\text{Benchmark}^i_{\text{MLLM}_{\text{subset}}}$ represents the performance of the MLLM trained on the subset on the $i$-th benchmark, and $\text{Benchmark}^i_{\text{MLLM}_{\text{full-set}}}$ denotes the performance of the MLLM trained on the full dataset on the $i$-th benchmark.
This metric quantifies the average relative accuracy of subset-trained MLLMs against full-set-trained MLLMs across all benchmarks, with a value closer to 100\% indicating that the subset achieves performance comparable to the full dataset.

\subsection{Baselines and Implementation Details}
\label{appendix:exp_baselines}
For the selection and implementation of baseline methods, we consider the following approaches and configurations.
\begin{itemize}
    \setlength{\itemsep}{0pt}
    \setlength{\parsep}{0pt}
    \setlength{\parskip}{0pt}

\item \textbf{EL2N and Perplexity: }
We first consider data selection methods based on EL2N~\cite{paul2021deep} and Perplexity~\cite{marion2023less}. 
These score-based methods are widely adopted as baselines for data selection in general visual instruction tuning datasets.
EL2N estimates sample quality via the Error L2-Norm score, defined as $\mathbb{E}\left[\|p(x) - y\|_2\right]$. 
Here, $p(\cdot)$ denotes the reference model, the $x$ denotes the input, and the $y$ denotes the label. 
This metric computes the average L2 distance between the model’s predictions and the labels of text tokens.
Perplexity measures the average negative log likelihood of next token prediction,  defined as $\exp\left(-\mathbb{E}\left[\log p(x)\right]\right)$. 
It evaluates the uncertainty present in the model’s predictions.
In addition, we follow prior research~\cite{marion2023less} to select samples with medium score values rather than top values, as it leads to better results.

\item \textbf{Coverage-based Coreset Selection (CCS): }
CCS~\cite{zheng2023coverage} aims to maximize coverage across the semantic space, ensuring an even distribution of both easy and difficult samples in the selected subset.
It divides a range of difficulty scores into equal-sized bins and randomly samples data from each bin while adhering to a uniform budget. 
Here we calculate the PPL scores of samples to quantify their relative ease or difficulty, and then conduct CCS-based sampling with the number of bins set to $50$ using these scores.

\item \textbf{COre INstruction Concept-skIll Data Election (COINCIDE): }
COINCIDE~\cite{lee2024concept} is a powerful method for data selection on general visual instruction tuning datasets, and once stood as the state of the art in this field. 
This method does not involve the calculation of gradients or explicit dependence on the task annotations of the dataset.
It first leverages the fixed TinyLLaVA-2B~\cite{zhou2024tinyllava} to extract both image and instruction features from the task-specific multiple layers of TinyLLaVA-2B, then concatenates all extracted features, performs spherical clustering on them, and selects samples from each cluster in proportion to the cluster’s overall transferability. 
Given that the chart comprehension capability of TinyLLaVA-2B is insufficient to extract valid information, we adapt COINCIDE to enable its effective application to capability gain probing on the ChartU training set. 
We extract features right after the multi-head self-attention module from layers $3, 10, 17, 24,$ and $31$ of LLaVA-Next-LLaMA3-8B, as well as layers $3, 9, 15, 21,$ and $27$ of Qwen2.5-VL-7B, since these layers meet COINCIDE’s requirement of covering multiple layers ranging from shallow to deep, and we fix the number of clusters at $1000$.
\end{itemize}

All data selection processes for the aforementioned methods are conducted on 4 RTX 4090 GPUs, and methods that involve clustering (including our \EXaMCaP) adopt Faiss-GPU~\cite{johnson2019billion} to accelerate clustering that requires a large number of clusters.
Additionally, we set the cluster number $L$ to 1000 and the size of random candidate set $m$ to 100.
For the kernel bandwidth parameter $\sigma$ of the Gaussian similarity measure, we set it to $0.5$ for all experiments. 
This choice is motivated by the naturally high similarity of intra-cluster samples, as the small $\sigma$ enhances the Gaussian kernel’s sensitivity to subtle distance variations between samples and facilitates finer-grained discrimination of similar samples.
Due to page constraints in the main text, we present the ablation study on $\sigma$ in the Appendix~\ref{appendix:exp_ablation}. 
We conduct all training with a learning rate of $1 \times 10^{-4}$, a batch size of 8 and LoRA~\cite{hu2022lora} configured with a rank of 8 and an alpha value of $16$, which are consistent with ECD’s~\cite{yang2025effective} configurations. 
All training is conducted on 4 RTX 4090 GPUs.

\section{More detailed experimental results}
\subsection{Details of Further Analysis Results}
\label{appendix:exp_futher}
In this subsection, we show more detailed experimental results of the experiments on Subset Size and alternative structures of MLLMs presented in Section~\ref{sec:further_analysis}.

Table~\ref{tab:ablation_DiffTrainSize_appendix} details the impact of subset sizes on ChartU dataset capability gain probing. 
The subset sizes range from 20K to 80K in increments of 20K, with 60K acting as the threshold for subsets accounting for approximately 20\% or less of the total training set size. 
Training on the full dataset takes approximately 58.1 hours, while training on the 60K and 80K subsets takes around 10.9 hours and 14.6 hours, respectively.

Table~\ref{tab:ablation_DiffEmbTrainDiffModel_appendix} presents more detailed experimental results regarding the impact of alternative data information extraction MLLMs and target MLLMs on capability gain probing for the ChartU dataset. 
We employ two models with distinct architectures and varying ChartU capabilities, LLaVA-Next-LLaMA3-8B and Qwen2.5-VL-7B.

\begin{table*}[]
\caption{
More detailed results on the impact of subset sizes on capability gains probing. 
}
\centering
\resizebox{\textwidth}{!}{
\begin{tabular}{clccccccccccc}
\toprule
& \multicolumn{1}{c}{} 
& \multicolumn{7}{c}{\textsc{Out-of-Domain}} 
&  
& \multicolumn{2}{c}{\textsc{In-Domain}} 
&  \\ 

\cmidrule(lr){3-9} \cmidrule(lr){11-12}

& \multicolumn{1}{c}{} 
& \multicolumn{2}{c}{CharXiv} 
& \multicolumn{2}{c}{ChartQA} 
& ChartX 
& \multicolumn{2}{c}{ReachQA} 
&  
& \multicolumn{2}{c}{$\text{ECD}\,_\text{Bench}$} 
&  \\

\cmidrule(lr){3-4} 
\cmidrule(lr){5-6} 
\cmidrule(lr){7-7} 
\cmidrule(lr){8-9}  
\cmidrule(lr){11-12} 

\multirow{-3}{*}{\textsc{
    \begin{tabular}[c]{@{}c@{}}Train Set\\ (Size)\end{tabular}
}} 
& \multicolumn{1}{c}{\multirow{-3}{*}{\textsc{Methods}}} 
& reas. 
& desc. 
& human 
& augmented 
& reas. 
& reas. 
& reco 
& \multirow{-3}{*}{
    \textsc{
        \begin{tabular}[c]{@{}c@{}}Avg-Rel\\ (\%)\end{tabular}
    }
} 
& reas. 
& desc. 
& \multirow{-3}{*}{
    \textsc{
        \begin{tabular}[c]{@{}c@{}}Avg-Rel\\ (\%)\end{tabular}
    }
} \\ 

\midrule

- & Original LLaVA-Next & 17.80 & 37.92 & 54.72 & 78.48 & 28.73 & 7.30 & 21.80 & - & 8.17 & 18.95 & - \\ 
\midrule  

321K & Full Set to Train & 22.80 & 57.28 & 61.68 & 78.00 & 50.69 & 15.50 & 40.20 & 100.00 & 23.04 & 51.96 & 100.00 \\ 

\midrule  

& Random & 19.50 & 53.49 & 58.20 & 79.52 & 40.63 & 8.30 & 28.20 & 82.72 & 12.58 & 33.66 & 77.61 \\

& Perplexity & 21.90 & 50.53 & 58.00 & 82.08 & 39.93 & 8.80 & 28.40 & 84.25 & 12.09 & 34.48 & 78.73 \\

& EL2N & 19.50 & 47.45 & 58.64 & 80.88 & 41.75 & 10.40 & 29.40 & 84.25 & 13.89 & 32.92 & 79.26 \\

& CCS & 22.00 & 54.10 & 56.32 & 81.04 & 38.54 & 8.90 & 27.60 & 84.04 & 12.50 & 34.31 & 78.73 \\

& COINCIDE & 21.30 & 51.09 & 56.88 & 81.44 & 39.50 & 10.00 & 28.00 & 84.48 & 13.32 & 33.42 & 79.27 \\

\multirow{-6}{*}{20k}   
& \cellcolor[HTML]{D6DCE4}\textbf{Ours} 
& \cellcolor[HTML]{D6DCE4}20.70 
& \cellcolor[HTML]{D6DCE4}49.86 
& \cellcolor[HTML]{D6DCE4}58.88 
& \cellcolor[HTML]{D6DCE4}81.92 
& \cellcolor[HTML]{D6DCE4}41.84 
& \cellcolor[HTML]{D6DCE4}10.00 
& \cellcolor[HTML]{D6DCE4}31.40 
& \cellcolor[HTML]{D6DCE4}86.21 
& \cellcolor[HTML]{D6DCE4}12.91 
& \cellcolor[HTML]{D6DCE4}33.50 
& \cellcolor[HTML]{D6DCE4}80.44 \\ 
\midrule

& Random & 21.70 & 54.45 & 57.20 & 80.40 & 43.32 & 10.20 & 31.40 & 87.92 & 15.36 & 37.01 & 83.70 \\

& Perplexity & 22.90 & 52.36 & 55.36 & 81.60 & 41.75 & 10.20 & 29.00 & 86.65 & 12.66 & 37.09 & 81.43 \\

& EL2N & 20.70 & 48.94 & 60.00 & 80.88 & 46.44 & 10.40 & 31.90 & 87.89 & 14.54 & 36.27 & 83.13 \\

& CCS & 21.70 & 54.77 & 59.04 & 80.88 & 43.06 & 9.00 & 28.90 & 86.44 & 14.05 & 37.66 & 82.06 \\

& COINCIDE & 21.10 & 54.99 & 57.92 & 82.00 & 42.80 & 10.60 & 32.20 & 88.64 & 14.62 & 37.09 & 83.93 \\

\multirow{-6}{*}{40k} 
& \cellcolor[HTML]{D6DCE4}\textbf{Ours} 
& \cellcolor[HTML]{D6DCE4}21.00 
& \cellcolor[HTML]{D6DCE4}52.55 
& \cellcolor[HTML]{D6DCE4}58.64 
& \cellcolor[HTML]{D6DCE4}80.16 
& \cellcolor[HTML]{D6DCE4}46.35 
& \cellcolor[HTML]{D6DCE4}10.60 
& \cellcolor[HTML]{D6DCE4}31.50 
& \cellcolor[HTML]{D6DCE4}88.55 
& \cellcolor[HTML]{D6DCE4}15.77 
& \cellcolor[HTML]{D6DCE4}38.40 & \cellcolor[HTML]{D6DCE4}84.69 \\ 
\midrule

& Random & 22.90 & 54.15 & 59.04 & 80.88 & 45.75 & 10.40 & 33.30 & 90.65 & 14.87 & 38.07 & 85.82 \\

& Perplexity & 22.70 & 54.32 & 56.56 & 79.04 & 41.49 & 10.00 & 30.20 & 86.99 & 15.11 & 41.42 & 83.80 \\

& EL2N & 21.90 & 51.69 & 61.20 & 80.88 & 46.35 & 11.90 & 34.90 & 92.03 & 16.18 & 40.03 & 87.94 \\

& CCS & 22.60 & 55.27 & 58.08 & 74.88 & 43.58 & 10.50 & 31.40 & 88.23 & 15.20 & 39.79 & 84.46 \\

& COINCIDE & 22.30 & 54.67 & 58.24 & 79.44 & 43.92 & 11.10 & 31.70 & 89.52 & 15.60 & 39.54 & 85.60 \\
 
\multirow{-6}{*}{60k} 
& \cellcolor[HTML]{D6DCE4}\textbf{Ours} 
& \cellcolor[HTML]{D6DCE4}23.10 
& \cellcolor[HTML]{D6DCE4}54.16 
& \cellcolor[HTML]{D6DCE4}60.00 
& \cellcolor[HTML]{D6DCE4}81.84 
& \cellcolor[HTML]{D6DCE4}44.79 
& \cellcolor[HTML]{D6DCE4}12.00 
& \cellcolor[HTML]{D6DCE4}34.80 
& \cellcolor[HTML]{D6DCE4}92.92 
& \cellcolor[HTML]{D6DCE4}18.14 
& \cellcolor[HTML]{D6DCE4}40.03 
& \cellcolor[HTML]{D6DCE4}89.58 \\ 
\midrule

& Random & 22.40 & 57.49 & 60.16 & 79.92 & 46.79 & 12.30 & 33.80 & 93.48 & 15.69 & 41.01 & 89.04 \\

& Perplexity & 23.50 & 55.05 & 57.04 & 79.60 & 43.14 & 11.40 & 33.50 & 90.81 & 15.11 & 43.06 & 87.13 \\

& EL2N & 21.00 & 51.65 & 60.72 & 80.96 & 48.60 & 12.40 & 35.70 & 92.74 & 16.91 & 42.24 & 89.32 \\

& CCS 
& 21.10 
& 56.30 
& 58.08 
& 76.00 
& 44.62 
& 10.80 
& 32.90 
& 88.85 
& 16.26 
& 42.16 
& 85.97 \\

& COINCIDE & 22.10 & 55.84 & 59.68 & 79.60 & 46.18 & 12.20 & 35.30 & 92.98 & 16.42 & 42.73 & 89.37 \\
 
\multirow{-6}{*}{80k} 
& \cellcolor[HTML]{D6DCE4}\textbf{Ours} 
& \cellcolor[HTML]{D6DCE4}22.50 
& \cellcolor[HTML]{D6DCE4}54.89 
& \cellcolor[HTML]{D6DCE4}61.60 
& \cellcolor[HTML]{D6DCE4}80.16 
& \cellcolor[HTML]{D6DCE4}48.00 
& \cellcolor[HTML]{D6DCE4}13.20 
& \cellcolor[HTML]{D6DCE4}34.10 
& \cellcolor[HTML]{D6DCE4}94.55 
& \cellcolor[HTML]{D6DCE4}17.08 
& \cellcolor[HTML]{D6DCE4}41.67 
& \cellcolor[HTML]{D6DCE4}90.68 \\ 
\bottomrule 
\end{tabular}
}

\label{tab:ablation_DiffTrainSize_appendix}
\end{table*}
\begin{table*}[]
\caption{
More detailed results regarding the impact of alternative extraction and target MLLMs on capability gains probing.
}
\centering
\resizebox{\textwidth}{!}{
\begin{tabular}{cclccccccccccc}
\toprule
&  
& \multicolumn{1}{c}{} 
& \multicolumn{7}{c}{\textsc{Out-of-Domain}} 
&  
& \multicolumn{2}{c}{\textsc{In-Domain}} 
&  \\ 

\cmidrule(lr){4-10} \cmidrule(lr){12-13}

&  
& \multicolumn{1}{c}{} 
& \multicolumn{2}{c}{charXiv} 
& \multicolumn{2}{c}{ChartQA} 
& ChartX 
& \multicolumn{2}{c}{ReachQA} 
&  
& \multicolumn{2}{c}{$\text{ECD}\,_\text{Bench}$} 
&  \\

\cmidrule(lr){4-5} 
\cmidrule(lr){6-7}
\cmidrule(lr){8-8}  
\cmidrule(lr){9-10}
\cmidrule(lr){12-13}

\multirow{-3}{*}{\textsc{\begin{tabular}[c]{@{}c@{}}Training\\ Models\end{tabular}}} 
& \multirow{-3}{*}{\textsc{\begin{tabular}[c]{@{}c@{}}Embedding\\ Models\end{tabular}}} 
& \multicolumn{1}{c}{\multirow{-3}{*}{\textsc{Methods}}} 
& reas. 
& desc. 
& human 
& augmented 
& reas. 
& reas. 
& reco. 
& \multirow{-3}{*}{\textsc{\begin{tabular}[c]{@{}c@{}}Avg-Rel\\ (\%)\end{tabular}}} 
& reas. 
& desc. 
& \multirow{-3}{*}{\textsc{\begin{tabular}[c]{@{}c@{}}Avg-Rel\\ (\%)\end{tabular}}} \\ 

\midrule

&  
& \cellcolor[HTML]{E7E6E6}Full Set 
& \cellcolor[HTML]{E7E6E6}22.80 
& \cellcolor[HTML]{E7E6E6}57.28 
& \cellcolor[HTML]{E7E6E6}61.68 
& \cellcolor[HTML]{E7E6E6}78.00 
& \cellcolor[HTML]{E7E6E6}50.69 
& \cellcolor[HTML]{E7E6E6}15.50 
& \cellcolor[HTML]{E7E6E6}40.20 
& \cellcolor[HTML]{E7E6E6}100.00
& \cellcolor[HTML]{E7E6E6}23.04 
& \cellcolor[HTML]{E7E6E6}51.96 
& \cellcolor[HTML]{E7E6E6}100.00\\ 
\cmidrule(lr){3-14}

& \multirow{-2}{*}{-} & Random 
& 22.90 & 54.15 & 59.04 & 80.88 & 45.75 & 10.40 & 33.30 & 90.65 & 14.87 & 38.07 & 85.82 \\ 
\cmidrule(lr){2-14}

&  & Perplexity 
& 22.70 & 54.32 & 56.56 & 79.04 & 41.49 & 10.00 & 30.20 & 86.99 & 15.11 & 41.42 & 83.80 \\
&  & EL2N 
& 21.90 & 51.69 & 61.20 & 80.88 & 46.35 & 11.90 & 34.90 & 92.03 & 16.18 & 40.03 & 87.94 \\
&  & CCS 
& 22.60 & 55.27 & 58.08 & 74.88 & 43.58 & 10.50 & 31.40 & 88.23 & 15.20 & 39.79 & 84.46 \\
&  & COINCIDE 
& 22.30 & 54.67 & 58.24 & 79.44 & 43.92 & 11.10 & 31.70 & 89.52 & 15.60 & 39.54 & 85.60 \\

& \multirow{-5}{*}{LLaVA-Next} & \cellcolor[HTML]{D6DCE4}Ours & \cellcolor[HTML]{D6DCE4}23.10 & \cellcolor[HTML]{D6DCE4}54.16 & \cellcolor[HTML]{D6DCE4}60.00& \cellcolor[HTML]{D6DCE4}81.84 & \cellcolor[HTML]{D6DCE4}44.79 & \cellcolor[HTML]{D6DCE4}12.00 & \cellcolor[HTML]{D6DCE4}34.80 & \cellcolor[HTML]{D6DCE4}92.92 & \cellcolor[HTML]{D6DCE4}18.14 & \cellcolor[HTML]{D6DCE4}40.03 & \cellcolor[HTML]{D6DCE4}89.58 \\ 
\cmidrule(lr){2-14}

&  & Perplexity & 22.00 & 49.26 & 59.92 & 80.00& 45.23 & 10.40 & 33.00 & 88.66 & 16.99 & 38.64 & 85.41 \\
&  & EL2N & 20.20 & 41.86 & 60.96 & 80.32 & 48.26 & 10.90 & 32.70 & 87.19 & 15.52 & 40.03 & 83.86 \\
&  & CCS & 21.40 & 53.62 & 56.72 & 80.16 & 43.49 & 9.40 & 34.30 & 87.71 & 14.54 & 39.87 & 83.76 \\

&  
& COINCIDE 
& 22.70
& 55.98
& 58.61 
& 81.36 
& 46.61 
& 10.90
& 32.20
& 91.29
& 16.34
& 41.18
& 87.69\\
\multirow{-12}{*}{LLaVA-Next} & \multirow{-5}{*}{Qwen2.5-VL} & \cellcolor[HTML]{D6DCE4}Ours & \cellcolor[HTML]{D6DCE4}22.70 & \cellcolor[HTML]{D6DCE4}53.44 & \cellcolor[HTML]{D6DCE4}61.60 & \cellcolor[HTML]{D6DCE4}80.08 & \cellcolor[HTML]{D6DCE4}46.18 & \cellcolor[HTML]{D6DCE4}12.20 & \cellcolor[HTML]{D6DCE4}33.50 & \cellcolor[HTML]{D6DCE4}92.65 & \cellcolor[HTML]{D6DCE4}16.83 & \cellcolor[HTML]{D6DCE4}39.22 & \cellcolor[HTML]{D6DCE4}88.56 \\ 
\midrule

&  & \cellcolor[HTML]{E7E6E6}Full Set 
& \cellcolor[HTML]{E7E6E6}38.60
& \cellcolor[HTML]{E7E6E6}74.35
& \cellcolor[HTML]{E7E6E6}80.48
& \cellcolor[HTML]{E7E6E6}90.24
& \cellcolor[HTML]{E7E6E6}71.27
& \cellcolor[HTML]{E7E6E6}39.70
& \cellcolor[HTML]{E7E6E6}68.80
& \cellcolor[HTML]{E7E6E6}100.00
& \cellcolor[HTML]{E7E6E6}41.99
& \cellcolor[HTML]{E7E6E6}70.34
& \cellcolor[HTML]{E7E6E6}100.00\\ 
\cmidrule(lr){3-14}

& \multirow{-2}{*}{-} 
& Random 
& 40.50 & 72.68 & 79.36 & 90.48 & 70.40 & 36.50 & 67.90 
& 98.71 
& 37.66 & 69.12 
& 97.66 \\ 
\cmidrule(lr){2-14}

&  
& Perplexity 
& 38.90 & 73.49 & 80.00 & 91.68 & 70.05 & 34.90 & 68.40 
& 98.03 
& 37.75 & 70.26 
& 97.34 \\

&  
& EL2N & 38.70 & 68.35 & 80.56 & 91.84 & 71.61 & 38.30 & 68.00
& 98.55 
& 39.54 & 69.77 
& 98.13 \\

&  
& CCS 
& 38.00 & 74.45 & 79.92 & 89.52 & 69.53 & 36.40 & 67.50 
& 97.78 
& 37.83 & 68.55 
& 96.89 \\

&  
& COINCIDE 
& 39.00
& 73.33
& 80.56
& 90.64
& 71.18
& 35.60
& 69.10
& 98.60
& 36.68
& 69.36
& 97.35 \\

& \multirow{-5}{*}{LLaVA-Next} 
& \cellcolor[HTML]{D6DCE4}Ours 
& \cellcolor[HTML]{D6DCE4}40.10 & \cellcolor[HTML]{D6DCE4}71.25 & \cellcolor[HTML]{D6DCE4}80.72 & \cellcolor[HTML]{D6DCE4}90.24 & \cellcolor[HTML]{D6DCE4}70.75 & \cellcolor[HTML]{D6DCE4}37.90 & \cellcolor[HTML]{D6DCE4}69.20 
& \cellcolor[HTML]{D6DCE4}99.33 
& \cellcolor[HTML]{D6DCE4}38.97 & \cellcolor[HTML]{D6DCE4}69.93 
& \cellcolor[HTML]{D6DCE4}98.62 \\ 
\cmidrule(lr){2-14}

&  & Perplexity & 37.40 & 63.98 & 81.28 & 92.00 & 71.27 & 38.00 & 68.90 & 97.39 & 40.60 & 68.22 & 97.27 \\
&  & EL2N & 38.80 & 64.63 & 80.16 & 92.96 & 71.53 & 38.50 & 70.10 & 98.47 & 38.50 & 69.36 & 97.73 \\
&  & CCS & 37.80 & 72.75 & 79.68 & 89.92 & 69.27 & 35.70 & 68.40 & 97.28 & 37.50 & 69.36 & 96.54 \\
&  
& COINCIDE 
& 41.90
& 73.05
& 80.88
& 90.96
& 70.83
& 35.80
& 68.60
& 99.62
& 38.48
& 69.28
& 98.86
\\
\multirow{-12}{*}{Qwen2.5-VL} & \multirow{-5}{*}{Qwen2.5-VL} & \cellcolor[HTML]{D6DCE4}Ours & \cellcolor[HTML]{D6DCE4}39.40 & \cellcolor[HTML]{D6DCE4}71.93 & \cellcolor[HTML]{D6DCE4}80.48 & \cellcolor[HTML]{D6DCE4}91.68 & \cellcolor[HTML]{D6DCE4}71.27 & \cellcolor[HTML]{D6DCE4}38.90 & \cellcolor[HTML]{D6DCE4}68.70 & \cellcolor[HTML]{D6DCE4}99.75 & \cellcolor[HTML]{D6DCE4}39.13 & \cellcolor[HTML]{D6DCE4}69.12 & \cellcolor[HTML]{D6DCE4}98.86 \\
\bottomrule
\end{tabular}
}

\label{tab:ablation_DiffEmbTrainDiffModel_appendix}
\end{table*}

\subsection{Details of Ablation Study}
\label{appendix:exp_ablation}

In this subsection, we present more detailed experimental results of the ablation studies.

For the Specific Design Choices and the Hyperparameter Settings presented in Section~\ref{sec:ablation}, the results are summarized in Tables~\ref{tab:ablation_components_appendix} and~\ref{tab:ablation_hyperparam_appendix}.

Due to page constraints of the main text, the ablation study on the kernel bandwidth parameter $\sigma$ of the Gaussian similarity measure is presented in Table~\ref{tab:ablation_sigma_appendix}.
We observe that \EXaMCaP achieves superior performance with $\sigma=0.5$. 
Furthermore, when $\sigma$ is set to a relatively small value, the performance on \textsc{Out-Of-Domain Avg-Rel} exhibits slight fluctuations across adjusted ranges of $\sigma$, and this finding validates the rationality of our design consideration. This consideration is grounded in the naturally high similarity of intra-cluster samples, as a small $\sigma$ enhances the Gaussian kernel’s sensitivity to subtle inter-sample distance variations and thus enables finer-grained differentiation of similar samples.

\section{Analysis of Visualization Within Clusters for Different Sampling Methods}
\label{appendix:exp_analysis_vis}

To further investigate the intra-cluster sampling methods discussed in Section~\ref{sec:ablation}, we conduct an extended visualization analysis comparing the entropy gain and sample distribution of \EXaMCaP against two baseline methods. 
We compute the entropy based on the definition provided in Equation~\ref{eq:von_neumann_entropy}, applying an exponential scaling to enhance the discriminability between different sampling outcomes. 
Specifically, to better illustrate the observed patterns in entropy gain, we select the ten largest clusters. 
Additionally, we quantify the percentage increase in the entropy of samples selected by \EXaMCaP and MMD$^2$-Minimize relative to that of samples selected by the Random baseline.

\begin{figure}[]
\centering
  \includegraphics[width=0.85\columnwidth]{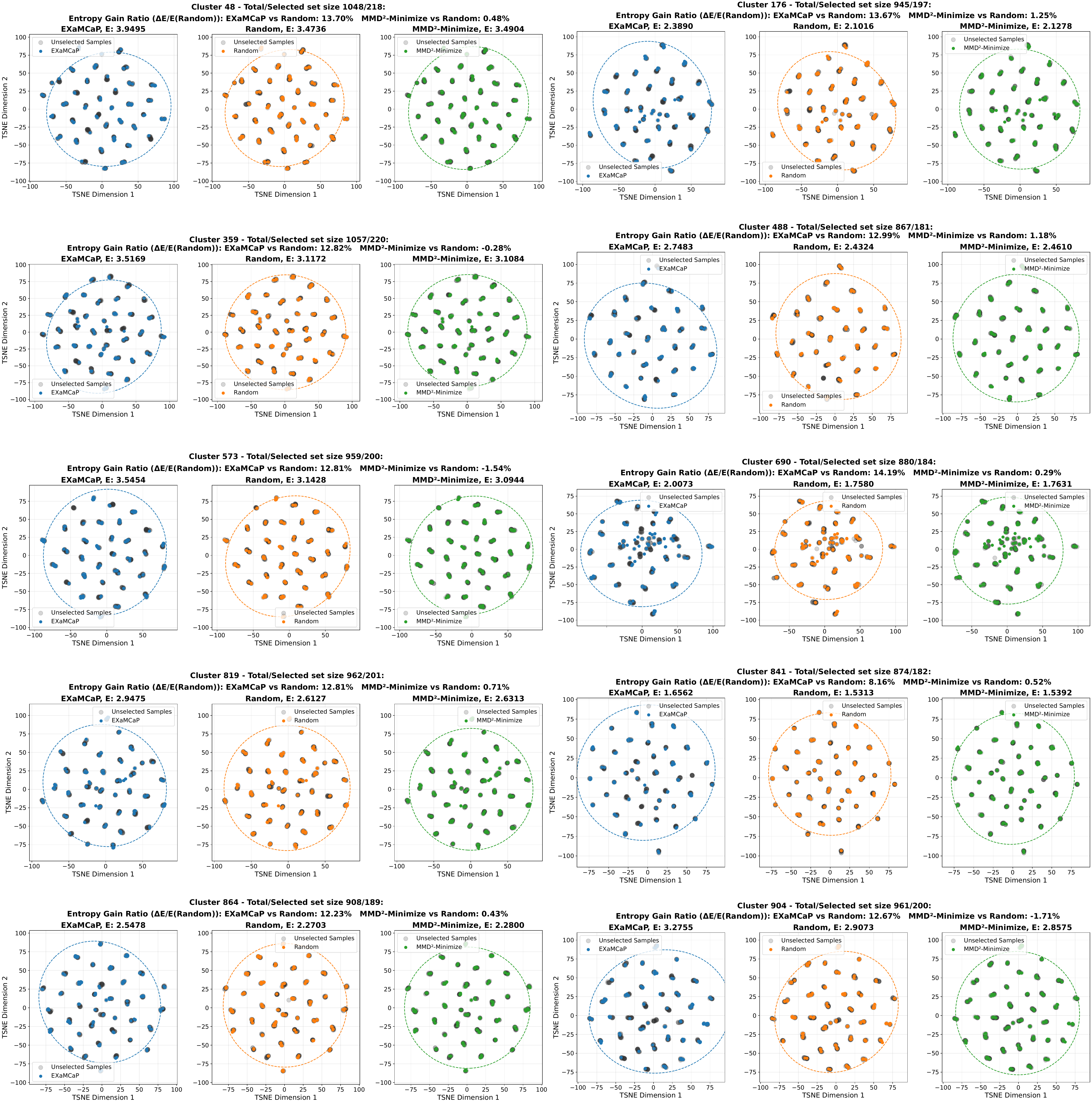}
   \caption{Distribution visualization and entropy gain comparison of samples selected by different intra-cluster sampling methods.}
  \label{fig:cluster_entropy_vis}
\end{figure}

Figure~\ref{fig:cluster_entropy_vis} presents the visualization results, where the dashed ellipses denote 95\% confidence ellipses of samples selected by each method. 
These ellipses are derived from the sample mean, covariance matrix, and eigenvalue decomposition to characterize the sample distribution in the 2D t-SNE space.
Combining visual observations and entropy gain ratio, \EXaMCaP achieves more diverse sampling within each cluster. 
The selected subsets cover the entire data space from core to peripheral regions, and well reflect the local density variations of the original data. 
This means that \EXaMCaP preserves the intrinsic structural features of the original distribution, without neglecting sparse but valuable peripheral samples. 
Moreover, \EXaMCaP achieves the highest and most stable entropy gain ratio against random sampling across all experiments, with the highest entropy among the three methods. 
These results confirm that subsets from \EXaMCaP contain rich information, thus maximizing sample diversity.
For the MMD$^2$-Minimize, its sampled points distribute uniformly in the t-SNE space, with significant distances between individual points and seemingly full coverage of the data space. 
However, its entropy is comparable to, or even lower than, that of random sampling. 
This is because MMD$^2$-Minimize aims to align the subset distribution with the original data. 
The method thus generates regular sampling patterns, which avoid random sampling but reduce information entropy. 
As a result, MMD$^2$-Minimize fails to maximize subset diversity despite its uniform spatial distribution.

\begin{table*}[!t]
\caption{More detailed ablation study results on specific design choices.
The row highlighted with \colorbox[HTML]{D6DCE4}{~~} corresponds to the unaltered \EXaMCaP results. 
The middle pane presents ablation results of \EXaMCaP regarding extreme sample filtering and cluster budget allocation strategies.
The last pane demonstrates the impacts of different intra-cluster sampling strategies on the performance of capability gains probing.}
\centering
\resizebox{\textwidth}{!}{
\begin{tabular}{cccccccccccccc}
\toprule
&  
&  
& \multicolumn{7}{c}{\textsc{Out-of-Domain}} 
&  
& \multicolumn{2}{c}{\textsc{In-Domain}} 
&  \\ 
\cmidrule(lr){4-10} 
\cmidrule(lr){12-13}

&  
&  
& \multicolumn{2}{c}{charXiv} 
& \multicolumn{2}{c}{chartqa} 
& ChartX 
& \multicolumn{2}{c}{reachqa} 
&  
& \multicolumn{2}{c}{ECD\_bench} 
&  \\

\cmidrule(lr){4-5} 
\cmidrule(lr){6-7} 
\cmidrule(lr){8-8} 
\cmidrule(lr){9-10} 
\cmidrule(lr){12-13} 

\multirow{-3}{*}{
    \textsc{\begin{tabular}[c]{@{}c@{}}Extreme\\ Sample\\ Filtering\end{tabular}}
} 
& \multirow{-3}{*}{
    \textsc{\begin{tabular}[c]{@{}c@{}}Cluster\\ Budget\\ Allocation\end{tabular}}
} 
& \multirow{-3}{*}{
    \textsc{\begin{tabular}[c]{@{}c@{}}Cluster\\ Sampling\\ Method\end{tabular}}
} 
& reas. 
& desc. 
& human 
& augmented 
& reas. 
& reas. 
& reco. 
& \multirow{-3}{*}{
\textsc{
    \begin{tabular}[c]{@{}c@{}}Avg-Rel\\ (\%)\end{tabular}
}
} 
& reas. 
& desc. 
& \multirow{-3}{*}{
\textsc{
    \begin{tabular}[c]{@{}c@{}}Avg-Rel\\ (\%)\end{tabular}
}
} \\ \midrule

\rowcolor[HTML]{F2F2F2} 
\multicolumn{3}{c}{Full Set to Train} & 22.80 & 57.28 & 61.68 & 78.00 & 50.69 & 15.50 & 40.20 & 100.00 & 23.04 & 51.96 & 100.00 \\ \midrule

\rowcolor[HTML]{D6DCE4}
\checkmark & Proportion & EXaM & 23.10 & 54.16 & 60.00 & 81.84 & 44.79 & 12.00 & 34.80 & 92.92 & 18.14 & 40.03 & 89.58 \\ 
\midrule

- & Proportion &  
& 22.60 & 52.67 & 60.64 & 80.24 & 46.09 & 12.50 & 34.30 & 92.74 & 17.08 & 40.11 & 88.94  \\

\checkmark & Average & \multirow{-2}{*}{EXaM} & 22.10 & 55.03 & 60.16 & 81.68 & 48.09 & 11.50 & 33.50 & 92.52 & 15.20 & 39.87 & 87.82 \\ 

\midrule

&  & Random & 21.90 & 54.75 & 59.68 & 80.08 & 44.88 & 10.80 & 33.30 & 90.30 & 15.60 & 40.93 & 86.51 \\
\multirow{-2}{*}{\checkmark} & 
\multirow{-2}{*}{Proportion} & MMD2-minimize & 23.70 & 54.18 & 58.56 & 80.16 & 43.58 & 10.40 & 31.40 & 89.63 & 15.11 & 40.60 & 85.68 \\

\bottomrule
\end{tabular}
}

\label{tab:ablation_components_appendix}
\end{table*}

\begin{table*}[!t]
\caption{More detailed experimental results on the impact of cluster numbers $L$ and random candidate set size $m$ on capability gains probing.
The row highlighted with \colorbox[HTML]{D6DCE4}{~~} corresponds to the results of \EXaMCaP with default settings. 
}
\centering
\resizebox{\textwidth}{!}{
\begin{tabular}{cccccccccccccc}
\toprule
& \multicolumn{2}{c}{} 
& \multicolumn{7}{c}{\textsc{Out-of-Domain}} 
&  
& \multicolumn{2}{c}{\textsc{In-Domain}} 
&  \\ 

\cmidrule(lr){4-10} \cmidrule(lr){12-13}

& \multicolumn{2}{c}{} 
& \multicolumn{2}{c}{charXiv} 
& \multicolumn{2}{c}{chartqa} 
& ChartX 
& \multicolumn{2}{c}{reachqa} 
&  
& \multicolumn{2}{c}{ECD\_bench} 
&  \\

\cmidrule(lr){4-5} 
\cmidrule(lr){6-7} 
\cmidrule(lr){8-8}  
\cmidrule(lr){9-10} 
\cmidrule(lr){12-13} 

\multirow{-3}{*}{\textsc{Methods}} & \multicolumn{2}{c}{\multirow{-3}{*}{\textsc{Hyperparameter}}} & reas. & desc. & human & augmented & reas. & reas. & reco. & \multirow{-3}{*}{
\textsc{
\begin{tabular}[c]{@{}c@{}}Avg-Rel\\ (\%)\end{tabular}}
} & reas. & desc. 
& \multirow{-3}{*}{
\textsc{
    \begin{tabular}[c]{@{}c@{}}Avg-Rel\\ (\%)\end{tabular}}
}\\ \midrule
\rowcolor[HTML]{F2F2F2} 
Full Set & - & - & 22.80 & 57.28 & 61.68 & 78.00 & 50.69 & 15.50 & 40.20 & 100.00 & 23.04 & 51.96 & 100.00 \\

\midrule

 &  & 500 & 23.60 & 54.43 & 59.12 & 79.84 & 44.62 & 11.90 & 34.30 & 92.41 & 16.50 & 40.11 & 88.41 \\
 
&  & \cellcolor[HTML]{D6DCE4}1000 & \cellcolor[HTML]{D6DCE4}23.10 & \cellcolor[HTML]{D6DCE4}54.16 & \cellcolor[HTML]{D6DCE4}60.00 & \cellcolor[HTML]{D6DCE4}81.84 & \cellcolor[HTML]{D6DCE4}44.79 & \cellcolor[HTML]{D6DCE4}12.00 & \cellcolor[HTML]{D6DCE4}34.80 & \cellcolor[HTML]{D6DCE4}92.92 & \cellcolor[HTML]{D6DCE4}18.14 & \cellcolor[HTML]{D6DCE4}40.03 & \cellcolor[HTML]{D6DCE4}89.58 \\

& \multirow{-3}{*}{Cluster Num.} & 1500 & 21.40 & 54.41 & 59.52 & 80.32 & 46.27 & 12.20 & 35.50 & 92.37 & 17.81 & 40.93 & 89.19 \\ \cmidrule(lr){2-14} 

&  & 10 
& 22.10 & 55.28 & 60.16 & 79.60 & 46.96 & 11.70 & 32.80 & 91.82 & 16.18 & 40.28 & 87.83 \\

&  & \cellcolor[HTML]{D6DCE4}100 & \cellcolor[HTML]{D6DCE4}23.10 & \cellcolor[HTML]{D6DCE4}54.16 & \cellcolor[HTML]{D6DCE4}60.00 & \cellcolor[HTML]{D6DCE4}81.84 & \cellcolor[HTML]{D6DCE4}44.79 & \cellcolor[HTML]{D6DCE4}12.00 & \cellcolor[HTML]{D6DCE4}34.80 & \cellcolor[HTML]{D6DCE4}92.92 & \cellcolor[HTML]{D6DCE4}18.14 & \cellcolor[HTML]{D6DCE4}40.03 & \cellcolor[HTML]{D6DCE4}89.58 \\

\multirow{-6}{*}{Ours} & \multirow{-3}{*}{Candidate Num.} & 1000 & 23.30 & 54.71 & 60.56 & 80.32 & 45.49 & 12.30 & 33.60 & 93.08 & 17.16 & 41.18 & 89.48 \\
\bottomrule
\end{tabular}
}

\label{tab:ablation_hyperparam_appendix}
\end{table*}
\begin{table*}[!t]
\caption{More detailed experimental results on the impact of kernel bandwidth parameter $\sigma$ on capability gains probing.
The row highlighted with \colorbox[HTML]{D6DCE4}{~~} corresponds to the results of \EXaMCaP with default settings. 
}
\centering
\resizebox{\textwidth}{!}{
\begin{tabular}{cccccccccccccc}
\toprule
& \multicolumn{2}{c}{} 
& \multicolumn{7}{c}{\textsc{Out-of-Domain}} 
&  
& \multicolumn{2}{c}{\textsc{In-Domain}} 
&  \\ 

\cmidrule(lr){4-10} \cmidrule(lr){12-13}

& \multicolumn{2}{c}{} 
& \multicolumn{2}{c}{charXiv} 
& \multicolumn{2}{c}{chartqa} 
& ChartX 
& \multicolumn{2}{c}{reachqa} 
&  
& \multicolumn{2}{c}{ECD\_bench} 
&  \\

\cmidrule(lr){4-5} 
\cmidrule(lr){6-7}
\cmidrule(lr){8-8}  
\cmidrule(lr){9-10}
\cmidrule(lr){12-13}

\multirow{-3}{*}{\textsc{Methods}} & \multicolumn{2}{c}{\multirow{-3}{*}{\textsc{
\begin{tabular}[c]{@{}c@{}}Kernel Bandwidth \\Parameter\end{tabular}
}}} & reas. & desc. & human & augmented & reas. & reas. & reco. & \multirow{-3}{*}{
\textsc{
\begin{tabular}[c]{@{}c@{}}Avg-Rel\\ (\%)\end{tabular}}
} & reas. & desc. 
& \multirow{-3}{*}{
\textsc{
    \begin{tabular}[c]{@{}c@{}}Avg-Rel\\ (\%)\end{tabular}}
}\\ \midrule
\rowcolor[HTML]{F2F2F2} 
Full Set & - & - & 22.80 & 57.28 & 61.68 & 78.00 & 50.69 & 15.50 & 40.20 & 100.00 & 23.04 & 51.96 & 100.00 \\

\midrule

 &  & $0.1$ 
& 22.70 
& 55.38 
& 61.92 
& 79.76 
& 47.05 
& 11.30 
& 33.40 
& 92.53 
& 16.34 
& 39.71 
& 88.34 \\
 
&  & \cellcolor[HTML]{D6DCE4}$0.5$ & \cellcolor[HTML]{D6DCE4}23.10 & \cellcolor[HTML]{D6DCE4}54.16 & \cellcolor[HTML]{D6DCE4}60.00 & \cellcolor[HTML]{D6DCE4}81.84 & \cellcolor[HTML]{D6DCE4}44.79 & \cellcolor[HTML]{D6DCE4}12.00 & \cellcolor[HTML]{D6DCE4}34.80 & \cellcolor[HTML]{D6DCE4}92.92 & \cellcolor[HTML]{D6DCE4}18.14 & \cellcolor[HTML]{D6DCE4}40.03 & \cellcolor[HTML]{D6DCE4}89.58 \\

\multirow{-3}{*}{Ours} & \multirow{-3}{*}{$\sigma$} & $1.0$ 
& 21.60 
& 54.11 
& 59.44 
& 79.95 
& 45.18 
& 12.80 
& 33.80 
& 91.98 
& 16.91 
& 40.85 
& 88.43 \\ 
\bottomrule
\end{tabular}
}

\label{tab:ablation_sigma_appendix}
\end{table*}

\section{Samples with Excessively High Perplexity Values}
\label{appendix:extreme_samples}

\begin{figure}[]
  \includegraphics[width=\columnwidth]{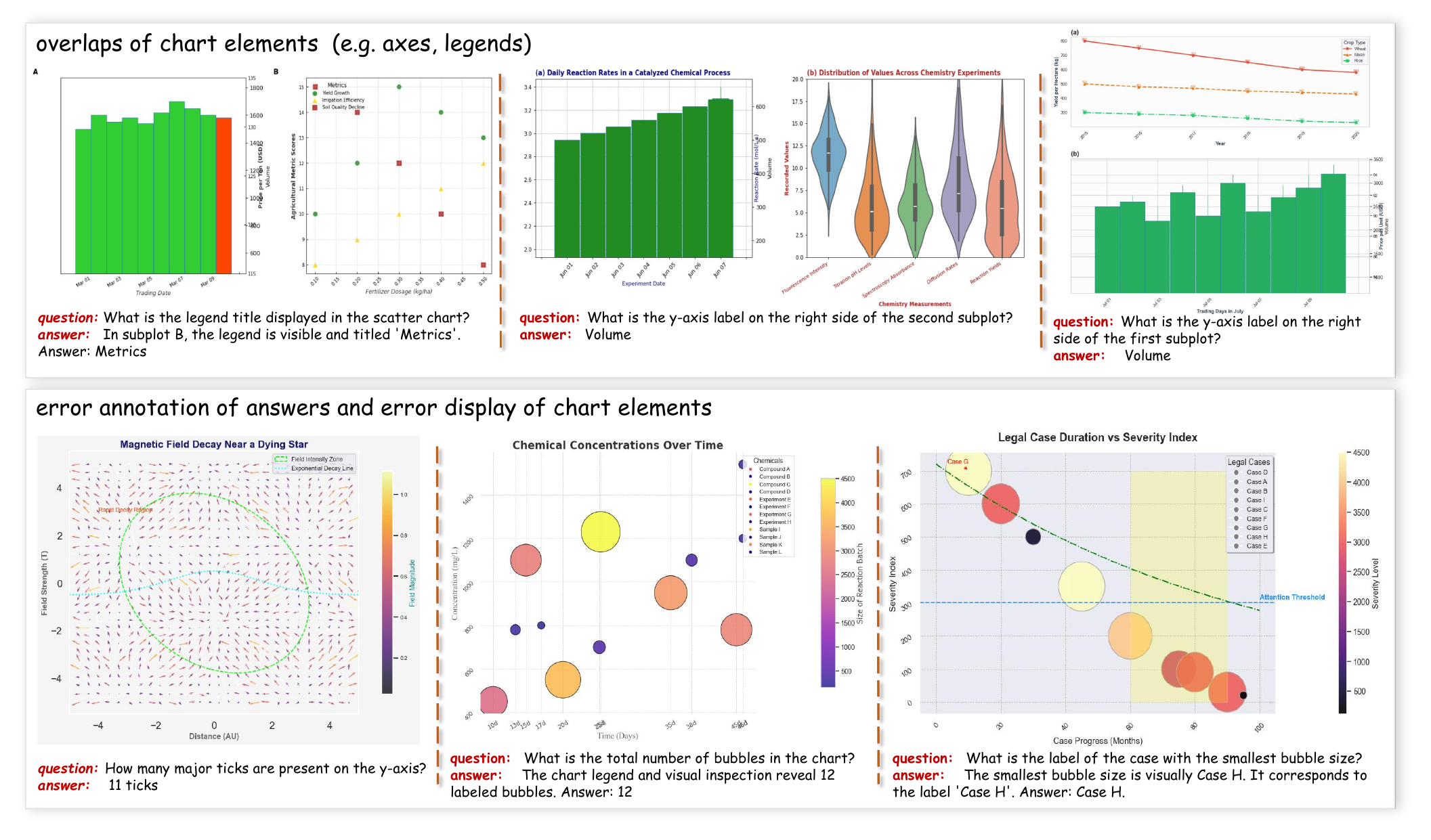}
   \caption{Some samples with excessively high perplexity values indeed suffer from issues such as explicit chart overlaps and annotation errors.}
  \label{fig:extreme_samples}
\end{figure}

We present some samples with relatively high perplexity (PPL) values (top 5\% of the highest PPL scores) in Figure~\ref{fig:extreme_samples}. 
These samples are derived from ECD, a training dataset that has undergone multiple rounds of quality filtering, and the PPL scores of the training data were obtained using the LLaVA-Next-LLaMA3 model.
As can be observed, high PPL values help identify low-quality samples, such as charts with overlapping chart elements, as well as samples with incorrect answer annotations and erroneous chart element displays. This observation is consistent with the findings of \citet{maharana2024adapt}.

\end{document}